\documentclass{article}

\PassOptionsToPackage{numbers}{natbib}
\usepackage[preprint]{neurips_2026}

\usepackage[utf8]{inputenc} 
\usepackage[T1]{fontenc}    
\usepackage{hyperref}       
\usepackage{url}            
\usepackage{booktabs}       
\usepackage{amsfonts}       
\usepackage{nicefrac}       
\usepackage{microtype}      
\usepackage{xcolor}         
\usepackage{natbib}
\usepackage{graphicx}
\usepackage{multirow}
\usepackage{algorithm}
\usepackage{algpseudocode}

\usepackage{makecell}
\usepackage{pifont}
\usepackage{array}
\usepackage{threeparttable}
\usepackage{adjustbox}
\usepackage{ragged2e}
\usepackage{tabularx}
\usepackage{svg}
\usepackage{wrapfig}
\usepackage{caption}
\usepackage{amsmath}
\usepackage{amssymb}

\newcolumntype{L}[1]{>{\raggedright\arraybackslash}p{#1}}
\newcolumntype{C}[1]{>{\centering\arraybackslash}p{#1}}

\algrenewcommand\algorithmicrequire{\textbf{Input:}}
\algrenewcommand\algorithmicensure{\textbf{Output:}}

\algnewcommand\algorithmichyperparameter{\textbf{Hyperparameter:}}
\algnewcommand\Hyperparameter{\item[\algorithmichyperparameter]}
\algnewcommand\algorithmicmodels{\textbf{Models:}}
\algnewcommand\Models{\item[\algorithmicmodels]}
\algnewcommand{\Initialize}{\State \textbf{Initialize }}

\algnewcommand\algorithmiclosses{\textbf{Losses:}}
\algnewcommand\Losses{\item[\algorithmiclosses]}
\algnewcommand\algorithmicoptimizers{\textbf{Optimizers:}}
\algnewcommand\Optimizers{\item[\algorithmicoptimizers]}

\title{DAD4TS: Data-Augmentation-Oriented Diffusion Model for Time-Series Forecasting with Small-Scale Data}

\author{%
  Masahiro Suzuki\thanks{Use footnote for providing further information
    about author (webpage, alternative address)---\emph{not} for acknowledging
    funding agencies.} \\
  Sony Group Corporation\\
  Tokyo, Japan
  \And
  Bohui Xia \\
  Sony Group Corporation \\
  Tokyo, Japan
  \AND
  Hiroto Yamamoto \\
  Sony Group Corporation \\
  Tokyo, Japan
  \And
  Masanori Miyahara \\
  Sony Group Corporation \\
  Tokyo, Japan
}

\begin{document}

\maketitle

\begin{abstract}

Small-scale data is a critical problem in time-series forecasting tasks. Data augmentation is an effective strategy for this task, but it has a limitation in generating meaningful data. To address this limitation, we propose DAD4TS, a diffusion-model-based data augmentation method with reinforcement learning, designed for time-series forecasting with small-scale data. In DAD4TS, a data generator is simultaneously trained with a time-series model and controlled by a reinforcement learning model to efficiently generate samples that improve the forecast accuracy of the time-series model. To support small-scale data, we use mathematical methods instead of conventional VAE methods to train the diffusion model by projecting the time-series data into the geometric space. We validated the effectiveness of DAD4TS with seven comparative methods through qualitative and quantitative experiments on six real-world datasets and eight time-series models. As a result, DAD4TS was validated on five datasets.
\end{abstract}




\section{Introduction}


Time-series forecasting (TSF) is one of the important tasks to solve a wide range of real-world problems, such as power management \citep{Liu_elec}, traffic forecasting \citep{li2018diffusion}, and weather forecasting \citep{dimri2020time}. On the other hand, there are many cases in which sufficient training data cannot be collected \citep{gonen2025time} in actual operation, and it is an important challenge to build effective forecasting models with small-scale data \citep{WANG2021504}. In addressing such cases, data augmentation is frequently employed as a strategy when training data is small scale \citep{wen2020time}, and various methods have been proposed to date, not limited to time-series domain \citep{chawla2002smote, Masato8852250, Jinsung_TimeGAN, desai2022TimeVAE, Masahiro10973545, yuan2024diffusionts, rahimi2025auggen}. Existing research in this field has extensively examined generation-based data augmentation \citep{Jinsung_TimeGAN, desai2022TimeVAE, Masahiro10973545, yuan2024diffusionts, yuan2025reaugmentmodelzooguidedrl, TarDiff, deng2025oats, YU2025112091, gonen2025time}, and it has been demonstrated that, when data augmentation methods are designed appropriately, such augmentation can indeed improve predictive performance \citep{Masahiro10973545, yuan2025reaugmentmodelzooguidedrl, TarDiff, deng2025oats, YU2025112091, gonen2025time, TAN2026152, dou2026autodatimeseries}. In this way, design approaches that take downstream tasks into account have been gaining attention in recent years.

\begin{table}
\centering
\caption{Average RMSE over multiple prediction lengths with input length fixed to 12. Results are averaged over forecast lengths $\in \{3,6,9,12\}$ for all datasets, except ILI where $\in \{2,4,8,12\}$. “Gaussian” refers to the case where time-series data—generated by adding Gaussian noise to the residuals from an STL decomposition of the original data and then reconstructing it—is used as training data alongside the original data. $\Delta$(\%) shows the relative change from baseline to “Gaussian”. ({\color{blue}blue} = degradation).}
\label{gussian_comparison}
\resizebox{\textwidth}{!}{
\begin{tabular}{llcccccccc|c}
\hline
Dataset & Methods & RNN & LSTM & Trans. & iTrans. & Patch. & Times. & S2IP & OLin. & Avg. \\
 & & \multicolumn{1}{c}{RMSE$\downarrow$} & \multicolumn{1}{c}{RMSE$\downarrow$} & \multicolumn{1}{c}{RMSE$\downarrow$} & \multicolumn{1}{c}{RMSE$\downarrow$} & \multicolumn{1}{c}{RMSE$\downarrow$} & \multicolumn{1}{c}{RMSE$\downarrow$} & \multicolumn{1}{c}{RMSE$\downarrow$} & \multicolumn{1}{c}{RMSE$\downarrow$} & \\
\hline
Employees & baseline & 1.32 & 1.38 & 1.30 & 0.153 & 0.144 & 0.192 & 0.184 & 0.201 & \\
 & +Gaussian & 1.30 & 1.42 & 1.41 & 0.161 & 0.148 & 0.207 & 0.182 & 0.209 & \\[3pt]
\cline{2-11}
\noalign{\vskip 1pt}
 & $\Delta$(\%) & -1.52\% & +2.72\% & +8.67\% & +5.30\% & +2.85\% & +7.77\% & -1.09\% & +3.85\% & {\color{blue}+3.57\%} \\
\hline
Forest & baseline & 1.15 & 1.27 & 1.08 & 1.02 & 1.10 & 1.08 & 1.02 & 1.03 & \\
 & +Gaussian & 1.18 & 1.28 & 1.14 & 1.02 & 1.07 & 1.10 & 1.05 & 1.05 & \\[3pt]
\cline{2-11}
\noalign{\vskip 1pt}
 & $\Delta$(\%) & +2.30\% & +0.988\% & +5.95\% & +0.220\% & -2.77\% & +1.60\% & +2.30\% & +1.26\% & {\color{blue}+1.48\%} \\
\hline
German & baseline & 1.20 & 1.44 & 1.34 & 1.34 & 1.47 & 1.37 & 1.48 & 1.39 & \\
 & +Gaussian & 1.25 & 1.47 & 1.37 & 1.41 & 1.49 & 1.42 & 1.43 & 1.53 & \\[3pt]
\cline{2-11}
\noalign{\vskip 1pt}
 & $\Delta$(\%) & +4.44\% & +2.07\% & +2.23\% & +5.22\% & +1.21\% & +3.71\% & -3.45\% & +9.91\% & {\color{blue}+3.17\%} \\
\hline
ILI & baseline & 1.83 & 1.81 & 1.98 & 0.567 & 0.580 & 0.584 & 0.601 & 0.614 & \\
 & +Gaussian & 1.84 & 1.87 & 1.99 & 0.572 & 0.549 & 0.567 & 0.555 & 0.595 & \\[3pt]
\cline{2-11}
\noalign{\vskip 1pt}
 & $\Delta$(\%) & +0.273\% & +3.31\% & +0.253\% & +0.793\% & -5.35\% & -2.91\% & -7.57\% & -3.05\% & -1.78\% \\
\hline
Inventories & baseline & 0.748 & 0.719 & 0.859 & 0.552 & 0.549 & 0.567 & 0.525 & 0.561 & \\
 & +Gaussian & 0.746 & 0.731 & 0.771 & 0.531 & 0.555 & 0.555 & 0.522 & 0.499 & \\[3pt]
\cline{2-11}
\noalign{\vskip 1pt}
 & $\Delta$(\%) & -0.267\% & +1.70\% & -10.2\% & -3.85\% & +1.19\% & -2.03\% & -0.524\% & -11.2\% & -3.14\% \\
\hline
Consumption & baseline & 4.72 & 4.58 & 4.51 & 0.342 & 0.398 & 0.394 & 0.688 & 0.530 & \\
 & +Gaussian & 4.71 & 4.52 & 4.52 & 0.380 & 0.479 & 0.388 & 0.469 & 0.498 & \\[3pt]
\cline{2-11}
\noalign{\vskip 1pt}
 & $\Delta$(\%) & -0.0530\% & -1.42\% & +0.222\% & +11.1\% & +20.4\% & -1.59\% & -31.9\% & -6.00\% & -1.14\% \\
\hline
\end{tabular}
}
\end{table}



\begin{wrapfigure}{r}{0.42\textwidth}
  \centering
  \captionsetup{skip=2pt}
  \begin{minipage}{0.85\linewidth}
    \centering
    \includegraphics[width=\linewidth]{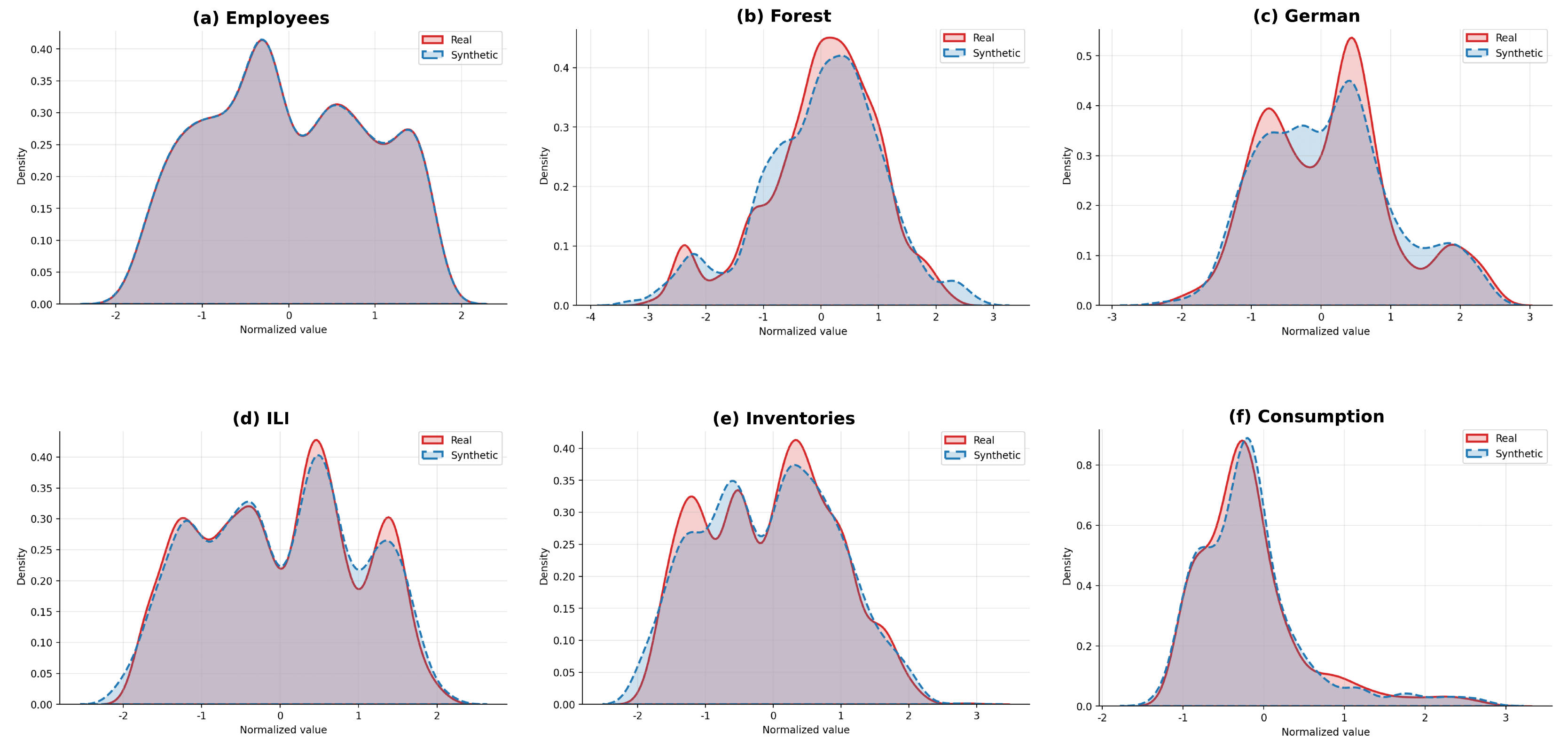}
    \caption{Distribution plots for each dataset of real and synthetic data.}
    \label{gaussian}
  \end{minipage}
\end{wrapfigure}

Despite this progress, many previous studies have focused on how closely the generated data resemble the original data, or have generated data to improve the accuracy of TSF models under the assumption that large amounts of data are available. However, when considering downstream tasks, generating and using data that resemble real data may also result in the generation and use of data that do not positively contribute to learning, which could adversely affect downstream tasks. In fact, as shown in Table\ref{gussian_comparison} based on preliminary experiments, it was confirmed in several cases that generating and using data resembling real data (Figure\ref{gaussian}) negatively affected downstream tasks. Furthermore, several previous studies have confirmed that the performance of generative models declines when the amount of data available for training is limited \citep{desai2022TimeVAE, gonen2025time, dou2026autodatimeseries}. Consequently, when training data is small scale, there are limitations to applying many of the previously reported studies to downstream tasks.


\setlength{\intextsep}{2pt}

On the other hand, many methods are static, with the data generation phase completed in a single step, and there are few methods that dynamically continue to generate samples optimized for TSF models. While the utility of synthetic samples depends on the current state of the forecasting model, even existing online augmentation frameworks provide only limited mechanisms for explicitly determining, based on forecasting performance, which samples should be generated and retained, and for integrating such decisions into the updates of both the generator and the forecaster.



In this paper, we introduce our method that Data-Augmentation-Oriented Diffusion Model for Time-Series Forecasting with Small-Scale Data (DAD4TS), a data augmentation method that generates samples to improve the accuracy of TSF models. The contributions of our method are as follows:

\begin{itemize}
\item \textbf{Optimization of Data Augmentation for Small-Scale Time-Series Data.} We propose a data augmentation framework for TSF models in small-scale data settings. The primary objective of our framework is to directly maximize downstream forecasting performance, and we demonstrate that it significantly improves time-series forecast accuracy under limited data settings compared to existing data augmentation methods.

\item \textbf{Value-Based Sampling of Generated Data.} Instead of indiscriminately using all generated samples for training, our framework introduces a Selector that continuously chooses informative generated samples based on a forecast-oriented reward. The Selector is trained to evaluate the utility of each generated sample by using improvements in the forecasting performance of the TSF models on a validation set as the reward signal.

\item \textbf{Joint training strategy.} Rather than treating the TSF models, the data generation model, and the Selector as independent components, we jointly train and update all three modules in an online manner. Data generation is not performed as a one-shot preprocessing step; instead, the Selector continuously determines which samples should be generated and retained based on their contribution to forecasting performance. These decisions are integrated into the updates of both the generator and the forecaster, resulting in a dynamic data augmentation process that continually adapts to the target forecasting task.

\item \textbf{TSF models-agnostic integration.} DAD4TS can be integrated in a manner independent of the architecture of TSF models and functions as a general-purpose extension module for time-series forecasting.
\end{itemize}

\section{Related Work}

\subsection{Generation of Time-Series Data}


Existing methods for time-series data generation include GAN and VAE based approaches \citep{Jinsung_TimeGAN, donahue2019wavegan, xu2020cot, desai2022TimeVAE}, as well as diffusion-model-based approaches \citep{Masahiro10973545, yuan2024diffusionts, shu2024data, zhang2025time, TarDiff, deng2025oats, YU2025112091, gonen2025time}. Similar to the image and speech domains, diffusion models have recently achieved remarkable success in time-series data generation. Building on the success of image generation using diffusion models, Suzuki et al. \citep{Masahiro10973545} visualized time-series data, used diffusion models to sample the data, and combined real and synthetic data to improve the accuracy of forecasting model. Yuan et al. \citep{yuan2025reaugmentmodelzooguidedrl} identify data points where the forecasting model is prone to overfitting, utilise these points as anchors, and dynamically perform data augmentation to prevent the model from overfitting to specific patterns, thereby improving forecast accuracy. Deng et al. \citep{deng2025oats} propose data augmentation for time-series foundation models (TSFMs), performing data augmentation that scales efficiently whilst calculating the balance of various time-series data, thereby improving the performance of TSFMs. Yu et al. \citep{YU2025112091} focus on temporal continuity—a factor previously overlooked in time-series data generation—and adopt a two-stage approach comprising a pre-training phase to preserve temporal continuity and a guidance augmentation phase to enable diffusion models to generate augmented data. Gonen et al. \citep{gonen2025time} propose an approach that involves training a diffusion model on a large volume of time-series data, followed by fine-tuning on a small dataset, thereby enabling efficient data generation for small-scale time-series data.

However, in many recent diffusion model approaches, data generation is performed only once, and data cannot be continuously generated toward the optimization of downstream tasks. In contrast, we aim to continuously perform data generation in conjunction with the training of TSF models, sequentially producing samples that contribute to improving forecasting accuracy. Moreover, many prior studies are based on DDPM \citep{ho2020denoising}, which requires a large number of sampling steps. Continuous data generation under such settings would incur prohibitive computational costs. Therefore, we use an approach that enables efficient data generation with a small number of sampling steps.

\subsection{Data selection methods}

The quality of training data used for machine learning and deep learning models has been extensively discussed in prior studies \citep{dvrl, Vitaly2020, just2023lava, jiang2023opendataval, Kaveen2025}. Since learning models indiscriminately learn from both informative data and noisy data, performance degradation can occur across various tasks regardless of how well the model itself is designed. To address this issue, Just et al. \citep{just2023lava} proposed a method for estimating data value that is independent of the downstream learning algorithm. Their approach evaluates data by comparing feature–label pairs between training and validation datasets.
In contrast, Yoon et al. \citep{dvrl} introduce a selection model that determines which data samples should be used for training. The selection model is trained such that a forecasting model trained on the selected data achieves higher validation performance than a model trained on the full dataset, thereby identifying valuable data samples. The DVRL method proposed by Yoon et al. \citep{dvrl} has been shown to be effective and highly generalizable across image, tabular, and language datasets.

As our study aims to continuously generate samples that contribute to improving the predictive performance of TSF models, it is necessary to control the diffusion model to enable effective data generation. In this context, the DVRL framework has the potential to serve as an effective mechanism. Accordingly, we apply the DVRL method to the time-series data generation task in our work.

\section{DAD4TS}

\begin{figure}
  \centering
    \includegraphics[
      width=0.7\linewidth,
    ]{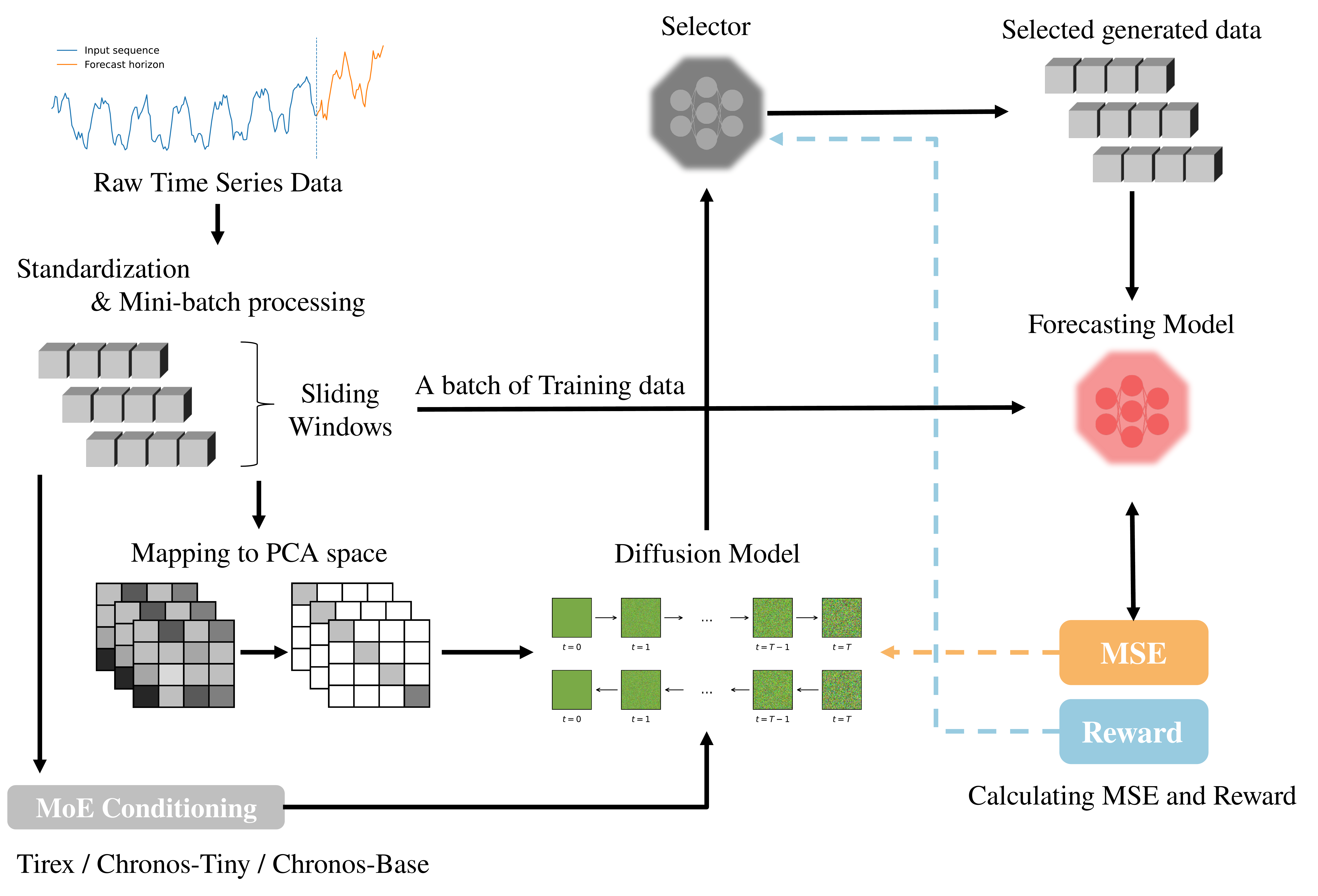}
  \caption{Overview of DAD4TS. After dividing the time-series data into batches, it is mapped onto a PCA space and fed into the diffusion model for training. The Selector then determines whether the data generated by the diffusion model is suitable for use; the generated data deemed suitable, together with the real data, is fed into the forecasting model for training. Finally, the forecasting model and the diffusion model are updated using MSE loss, and the Selector is updated using the calculated reward.}
  \label{overall}
\end{figure}


This section describes the DAD4TS. The proposed framework consists of two main components. The first is a Rectified Flow–based diffusion model, which enables data generation with a small number of sampling steps during the generation process. The second is Data Valuation using Deep Reinforcement Learning for Generated Samples (DVRL-G), a method that evaluates the value of samples generated by the generative model and determines whether they should be used for training the time-series forecasting model. Each of the following subsections details these components. Figure~\ref{overall} shows an overview of DAD4TS.

\subsection{Diffusion Model}
\label{sssec:diffusion_dad4ts}




This subsection describes the diffusion model developed in this study. Conventionally, diffusion models first map input data into a latent space using a VAE \citep{kingma2013auto}, and then learn a diffusion process on the latent representations to generate data \citep{rombach2022high, chen2025dove}. However, when the amount of available training data is limited, VAEs based on deep learning may not be able to learn sufficiently, making it difficult to capture the essential features of the data.

To address this issue, our approach maps time-series data into a geometric space without relying on deep learning–based methods. Specifically, the following steps are applied to the time-series data.

\begin{enumerate}
\item For a univariate time-series $\{d_t\}_{t=1}^{T}$, where $d_t \in \mathbb{R}$ for each $t$. We split the data into
training, validation, and test sets; $\mathcal{D}_{\mathrm{train}}$, $\mathcal{D}_{\mathrm{val}}$, and $\mathcal{D}_{\mathrm{test}}$, respectively. The split is in a ratio of 6:2:2. All splits are standardized to zero mean and unit variance using the mean and standard deviation estimated from $\mathcal{D}_{\mathrm{train}}$.
\item Prepare a dataset $\mathcal{D}_{\mathrm{train}}^{\mathrm{win}}$ from $\mathcal{D}_{\mathrm{train}}$ with a sliding window size of 1 and a batch size of B. The same settings are used for time-series forecasting.
\item Extract samples from $\mathcal{D}_{\mathrm{train}}^{\mathrm{win}}$ and calculate Gram matrices for each extracted sample to capture time-series features.
\item Given a mini-batch from $\mathcal{D}_{\mathrm{train}}^{\mathrm{win}}$, we fit a PCA \citep{MACKIEWICZ1993303} model per mini-batch on its representations and transform them into 2D embeddings.
\item For each two-dimensional vector obtained from each mini-batch in Step 4, we compute its outer product and set all off-diagonal elements to zero. This processing takes into account the correspondence of coordinates in the geometric space.
\end{enumerate}

The matrix computed through the above procedures constitutes the geometric space used by the diffusion model and is provided as input to the model. In addition, the time variable $t \in [0, 1]$ and a conditioning variable $c$ are simultaneously fed into the diffusion model. The conditioning variable (c) consists of embeddings extracted from TSFMs—specifically, Chronos-Tiny and Chronos-Base based on Chronos-Bolt \citep{ansari2024chronos}, as well as Tirex \citep{auer2025tirex}—which encode representations of the input time-series data. Among these embeddings, those selected by a Top-1 mixture-of-experts (MoE) setting are used as the conditioning input.

The diffusion model is implemented using a U-Net architecture, in which the conditioning variable $c$ and the time variable $t$ are integrated with the PCA-based representations via cross-attention.

\subsubsection{Training}
\label{sssec:training_dad4ts}
In training the diffusion model $\mathcal{M}_{\psi}$, we first learn representations in the geometric space. Subsequently, we compute the Mean Squared Error (MSE) loss between the predictions $\hat{\mathcal{Y}}_s$ of the TSF model $\mathcal{F}_{\theta}$ for the selected samples $\mathcal{X}_s$, which are chosen by the Selector described later in Section 3.2, and the corresponding selected samples $\mathcal{Y}_s$, and update $\mathcal{M}_{\psi}$ accordingly. Through this two-stage learning method, the diffusion model learns the geometric features of the time-series data and how the $\mathcal{F}_{\theta}$ model interprets the time-series data. 

In the initial training stage, the MSE loss is computed between the input data $x$ and the initial noise $x_t$, which has the same dimensionality as $x$ at time $t \in [0,1]$, and learning is performed based on the resulting MSE loss. This formulation is derived from Eq.~(\ref{eq:rectified_flow}), which is introduced later.
\begin{equation}
  v = \mathcal{M}_{\psi}(x_t, t,c),
  \ \, \mathrm{with } \,\, x_t=tx + (1-t)x_0
\end{equation}
\begin{equation}
  Loss = \frac{1}{n}\sum_{i=1}^{n} ((x-x_0) - v)^2
\end{equation}
$v$ is the expected velocity output by the diffusion model.
Furthermore, to incorporate condition c into the training process, we applied Classifier-Free Guidance (CFG) \citep{ho2021classifierfree}. In this study, 10\% of the training is conducted without conditions, enabling the model to learn to emphasise features when conditions are present. For the condition-free state, we provide the diffusion model with an empty condition 0. In other words, the following equation is given:
\begin{equation}
  v = \mathcal{M}_{\psi}(x_t, t, \emptyset)
\end{equation}
We optimize the diffusion model parameters using AdamW with a learning rate of $3\times10^{-3}$ and a weight decay of $0.1$.

\subsubsection{Rectified Flow}


When applying diffusion models to time-series data generation, approaches based on DDPM \citep{ho2020denoising} are commonly employed \citep{kong2020diffwave, Masahiro10973545, yuan2024diffusionts}. While the computational cost is not a major concern when data generation is performed only once, it becomes non-negligible when repeated generation is required, in which case the number of sampling steps emerges as a critical computational bottleneck. To address this issue, we adopt Rectified Flow \citep{liu2022flow}, an approach that enables data generation with diffusion models using a small number of sampling steps.

Rectified Flow is formulated as an Ordinary Differential Equation (ODE) model that transports a distribution $\pi_0$ to $\pi_1$ by following a path that is as close as possible to a straight line between noise and clean data. In this study, the number of sampling steps was set to 20.

Given $X_0 \sim \pi_0$ and $X_1 \sim \pi_1$, the Rectified Flow induced by the pair ($X_0, X_1$) is formulated as the ODE over time $t \in [0, 1]$. Let $Z_t$ be the values sampled from the distribution $\pi_t$. The dynamics are given by
\begin{equation}
  dZ_t=v(Z_t, t)dt
\end{equation}
which describes an ODE that transforms $Z_0$ into $Z_1$.

In addition, unlike conventional diffusion models, Rectified Flow generates data by solving a problem that estimates the expected velocity $v^{X}$ of a continuously differentiable stochastic process \{$X_t : t \in [0, 1]$\}, where the process is encouraged to follow, as closely as possible, straight-line trajectories from $X_0$ to $X_1$. Specifically, this requires computing
\begin{equation}
  v^{X}(x,t) = \mathbb{E}\!\left[ \dot{X}_t \mid X_t = x \right],
  \ \forall x \in \operatorname{supp}(X_t)
\end{equation}
\begin{equation}
\label{eq:rectified_flow}
  \min_{v}\int_{0}^{1} \mathbb{E}\left[ \left\| (X_1 - X_0) - v(X_t, t) \right\|^2 \right] \, dt,
  \ \, \mathrm{with } \, X_t = tX_1+(1-t)X_0.
\end{equation}

\subsubsection{Sampling}
\label{sssec:sampling_dad4ts}

\setlength{\intextsep}{4pt}

\begin{wrapfigure}{r}{0.5\textwidth}
  \centering
  \begin{minipage}{\linewidth}
    \begin{algorithm}[H]
    \caption{Sampling Process}
    \label{alg:sampling}
    \begin{algorithmic}[1]
    \Require Time $t_0 \in [0,1]$, $t_1 \in [0,1]$ $(t_0 < t_1)$, initial noise $x_0$, sampling step $\mathcal{T}$, conditioning variable $c$
    \Models Diffusion model $\mathcal{M}_{\psi}$ \textbf{Initialize} $\psi$
    \Ensure Sampling data $\hat{X}$

    \State $h \gets t_1 - t_0$
    \State $v_0 \gets \mathcal{M}_{\psi}(x_0, t_0, c)$
    \State $x_\epsilon \gets x_0+hv_0$
    \State $v_\epsilon \gets \mathcal{M}_{\psi}(x_\epsilon, t_1, c)$
    \State $x \gets x_0+\frac{h}{2}(v_0+v_\epsilon)$

    \For{$k = 1$ to $\mathcal{T}$}
        \State $\tilde{h} \gets t_{k+1}-t_k$
        \State $v_{k+1} \gets x+\frac{3}{2}\tilde{h}v_\epsilon -\frac{1}{2}\tilde{h}v_0$
        \State $\tilde{v}_{k+1} \gets x+\frac{1}{2}\tilde{h}(v_\epsilon+\mathcal{M}_{\psi}(v_{k+1}, t_{k+1}, c))$
        \State $v_0 \gets v_\epsilon$
        \State $v_\epsilon \gets \mathcal{M}_{\psi}(v_{k+1}, t_{k+1}, c)$
        \State $x \gets \tilde{v}_{k+1}$
     \EndFor

    \State $\hat{X} \gets x$
    \end{algorithmic}
    \end{algorithm}
  \end{minipage}
\end{wrapfigure}


In this study, the Heun method and the Adams-Bashforth and Adams–Moulton methods were applied to ODEs as samplers. The Heun method is a technique that produces higher quality results than the Euler method, which has traditionally been widely used as a sampler. The Adams–Bashforth and Moulton methods are linear multistep methods, which are commonly employed to obtain numerical approximations of ODE solutions. Unlike single-step methods such as the Euler method, the Adams–Bashforth method computes the numerical approximation at step $k$ by retaining and referencing information from previous steps up to $k-1$. The Adams-Moulton method is used to correct the predictions derived using the Adams-Bashforth method. However, Adams methods cannot be applied at the first sampling step. Therefore, the Heun method is employed to compute the approximate solution at the initial sampling step. In this study, second-order Adams methods are adopted. The overall sampling procedure is shown in Algorithm ~\ref{alg:sampling}.






Since CFG is applied in this study, both conditional and unconditional data are generated during the sampling process. Based on $\hat{X}$ computed in Algorithm ~\ref{alg:sampling}, the sampling procedure that incorporates both cases is given by
\begin{equation}
  \tilde{X} = (1+w)\hat{X}_c-w\hat{X}_\emptyset, 
  \ \, \mathrm{with } \, w=1.
\end{equation} 
After that, $\tilde{X}$ undergoes a process that reverses the transformation performed in Section \ref{sssec:diffusion_dad4ts}: first, Singular Value Decomposition (SVD) is applied to convert it into a two-dimensional vector; then, the inverse PCA transformation is performed to convert it into a Gram matrix; finally, the square roots of the values on the diagonal are taken to convert it back into the original time-series data.

\subsection{DVRL-G}
\subsubsection{Selector}
\label{sssec:selector_model}


In the DVRL-G framework, we first introduce a Selector $\mathcal{S}_{\phi}$ that selects generated data used for training the TSF model $\mathcal{F}_{\theta}$. The Selector $\mathcal{S}_{\phi}$ is implemented as a Transformer-based model. As inputs to $\mathcal{S}_{\phi}$, we use the data $\tilde{X}$ generated by the diffusion model, which are split into an input segment $x_b$ and a forecast segment $y_b$ according to the input length and forecast length required by $\mathcal{F}_{\theta}$. In addition, we provide the MSE loss between the ground-truth $y_b$ and the forecast $\hat{y}_b$ obtained by feeding $x_b$ into $\mathcal{F}_{\theta}$. These inputs enable the Selector to capture both the generated data and how $\mathcal{F}_{\theta}$ interprets each generated sample, allowing it to assess the value of individual generated data points. Based on these inputs, $\mathcal{S}_{\phi}$ outputs a selection vector in the form of weights $W_{vec}$, which represent the value of each sample. This process is formulated as follows:
\begin{equation}
  W_{vec}=\mathcal{S}_{\phi}(x_b, y_b, loss_b)
\end{equation}



Next, we prepare a Sampler that maps the vector output by $\mathcal{S}_{\phi}$ to the range [0, 1]. This Sampler follows a Bernoulli distribution. The generated data selected by this Sampler is used to train $\mathcal{F}_{\theta}$.

\subsubsection{Training}


For learning $\mathcal{S}_{\phi}$, we use the datasets $\mathcal{D}_{\mathrm{train}}^{\mathrm{win}}$ and $\mathcal{D}_{val}$ prepared in Section \ref{sssec:diffusion_dad4ts}. In addition, we construct $\mathcal{D}_{val}^{1/2}$, which consists of only the latter half of the validation data. First, starting from the $\mathcal{F}_{\theta}$, we train it for one epoch on $\mathcal{D}_{\mathrm{train}}^{\mathrm{win}}$ and compute the MSE loss $L_{base}$ on $\mathcal{D}_{val}^{1/2}$. Subsequently, $\mathcal{F}_{\theta}$ is trained using the generated data selected by the Selector together with a batch of training samples drawn from $\mathcal{D}_{\mathrm{train}}^{\mathrm{win}}$. After each batch update of $\mathcal{F}_{\theta}$, we compute the MSE loss $L_{val^{1/2}}$ on $\mathcal{D}_{val}^{1/2}$. The reward $\mathcal{R}$ is defined as $L_{base} - L_{val^{1/2}}$, and $\mathcal{S}_{\phi}$ is trained and updated so as to maximize $\mathcal{R}$.

To incorporate the influence of the selected samples into $\mathcal{R}$, we use the MSE loss $L_f$ obtained by feeding each generated sample into $\mathcal{F}_{\theta}$, as well as the log-probability under a Bernoulli distribution. When computing $L_f$, the loss is calculated for all generated samples regardless of whether they are selected by the Sampler. The loss $L_f$ reflects how $\mathcal{F}_{\theta}$ perceives the generated data (i.e., whether the data are easy or difficult to predict), while the log-probability indicates whether selected samples are valuable for training $\mathcal{F}_{\theta}$ and whether unselected samples are not valuable. Algorithm ~\ref{alg:dvrlg_process} (in the Appendix~\ref{DVRLG_Algorithm}) shows the sequence of steps in the DVRL-G process. The entire training process of DAD4TS is DVRL-G itself.
We optimize the Selector parameters with Adam, using a learning rate of $1\times10^{-3}$.

\section{Evaluation Experiment}
In this section, we describe TSF models and the comparison methods. Regarding the dataset we used, the evaluation metrics and the execution environment are described in the Appendix~\ref{dataset},~\ref{metrics},~\ref{environment}.

\subsection{TSF Models}

In previous studies, evaluations often relied on only a few time-series forecasting models \citep{Jinsung_TimeGAN, desai2022TimeVAE, yuan2024diffusionts, yuan2025reaugmentmodelzooguidedrl, gonen2025time, TAN2026152, dou2026autodatimeseries}, which carried the risk of producing biased results. Therefore, in this paper, we conducted an evaluation using eight TSF models, ranging from conventional to state-of-the-art models. The time-series forecasting models include RNN, LSTM \citep{hochreiter1997long}, Transformer \citep{vaswani2017attention}, iTransformer \citep{liu2023itransformer}, PatchTST \citep{nie2022time}, TimesNet \citep{wu2022timesnet}, S$^2$IP-LLM \citep{pan2024s}, and OLinear \citep{yue2025olinear}.

\subsection{Comparison Methods}

To evaluate the proposed method, we selected TimeGAN \citep{Jinsung_TimeGAN}, TimeVAE \citep{desai2022TimeVAE}, SMFG \citep{Masahiro10973545}, Diffusion-TS \citep{yuan2024diffusionts}, ReAugment \citep{yuan2025reaugmentmodelzooguidedrl}, DiffAT \citep{YU2025112091}, and AutoDA-Timeseries \citep{dou2026autodatimeseries} from previous studies as comparison methods.
These methods selected include data augmentation methods that improve the accuracy of downstream tasks despite limited training data \citep{Masahiro10973545}, methods that generate data closely resembling the distribution of the training data \citep{Jinsung_TimeGAN, desai2022TimeVAE, yuan2024diffusionts}, methods that use reinforcement learning to generate data for downstream tasks \citep{yuan2025reaugmentmodelzooguidedrl}, and methods that perform sequential data augmentation \citep{dou2026autodatimeseries}. By comparing DAD4TS with these methods, we evaluate its ability to optimize data augmentation for small-scale time-series data.

\section{Experimental Results}

\setlength{\tabcolsep}{1pt}
\renewcommand{\arraystretch}{0.60}
\begin{table}[t]
\centering
\caption{Average RMSE and DTW over multiple forecast lengths with input length fixed to 12. Results are averaged over forecast lengths $\in\{3,6,9,12\}$ for all datasets, except ILI where $\in\{2,4,8,12\}$. R represents RMSE, and D represents DTW. '-' indicates that the experiment could not be conducted in our experimental environment.}
\label{all_avg4pred_rmse}
\resizebox{\linewidth}{!}{%
\begin{tabular}{@{} l l r r *{16}{c} @{}}
\hline
Dataset & Methods & \multicolumn{2}{c}{Imp.(\%)} & \multicolumn{2}{c}{RNN} & \multicolumn{2}{c}{LSTM} & \multicolumn{2}{c}{Trans.} & \multicolumn{2}{c}{iTrans.} & \multicolumn{2}{c}{Patch.} & \multicolumn{2}{c}{Times.} & \multicolumn{2}{c}{S2IP.} & \multicolumn{2}{c}{OLin.} \\
 &  & Rate $\uparrow$ & Mean $\downarrow$ & R $\downarrow$ & D $\downarrow$ & R $\downarrow$ & D $\downarrow$ & R $\downarrow$ & D $\downarrow$ & R $\downarrow$ & D $\downarrow$ & R $\downarrow$ & D $\downarrow$ & R $\downarrow$ & D $\downarrow$ & R $\downarrow$ & D $\downarrow$ & R $\downarrow$ & D $\downarrow$  \\
\hline
Employees & baseline & 0.00 & 0.00 & 1.32 & 10.1 & 1.38 & 10.5 & 1.30 & 9.67 & 0.153 & 1.12 & 0.144 & 0.953 & 0.192 & 1.49 & 0.184 & 1.31 & 0.201 & 1.47 \\
 & TimeGAN & 48.4 & -2.60 & 1.30 & 9.89 & 1.45 & 10.9 & 1.43 & 10.8 & 0.150 & 1.05 & 0.142 & 0.948 & 0.169 & 1.23 & 0.182 & 1.34 & 0.148 & 0.942 \\
 & TimeVAE & 48.4 & 0.269 & 1.34 & 10.4 & 1.33 & 10.0 & 1.28 & 9.55 & 0.169 & 1.25 & 0.147 & 0.986 & 0.175 & 1.32 & 0.202 & 1.48 & 0.184 & 1.30 \\
 & SMFG & 59.4 & -13.6 & 1.36 & 10.6 & 1.55 & 11.8 & 1.54 & 11.7 & \textbf{\textcolor{red}{0.115}} & \textbf{\textcolor{red}{0.734}} & \textbf{\textcolor{red}{0.117}} & \textbf{\textcolor{red}{0.725}} & \textcolor{blue}{0.132} & \textbf{\textcolor{red}{0.839}} & \textbf{\textcolor{red}{0.114}} & \textbf{\textcolor{red}{0.716}} & \textbf{\textcolor{red}{0.131}} & \textbf{\textcolor{red}{0.855}} \\
 & Diffusion-TS & 64.1 & -9.20 & 1.38 & 10.8 & 1.41 & 10.8 & 1.69 & 12.8 & \textcolor{blue}{0.130} & \textcolor{blue}{0.881} & \textcolor{blue}{0.126} & \textcolor{blue}{0.793} & \textbf{\textcolor{red}{0.130}} & \textcolor{blue}{0.860} & \textcolor{blue}{0.130} & \textcolor{blue}{0.856} & 0.154 & 1.17 \\
 & ReAugment & 54.7 & -5.29 & 1.31 & 9.88 & 1.41 & 10.7 & 1.37 & 10.1 & 0.150 & 1.00 & 0.146 & 0.963 & 0.151 & 1.07 & 0.159 & 1.03 & 0.165 & 1.06 \\
 & DiffAT & 54.7 & 0.0758 & 1.19 & 8.85 & 1.43 & 10.6 & 1.44 & 10.8 & 0.184 & 1.34 & 0.140 & 0.936 & 0.158 & 1.12 & 0.177 & 1.21 & 0.195 & 1.39 \\
 & AutoDA & \textcolor{blue}{81.2} & \textcolor{blue}{-28.1} & \textcolor{blue}{0.588} & \textcolor{blue}{4.00} & \textcolor{blue}{0.453} & \textcolor{blue}{3.03} & \textcolor{blue}{0.799} & \textcolor{blue}{4.61} & 0.149 & 0.985 & 0.136 & 0.846 & 0.151 & 1.03 & 0.148 & 0.987 & 0.162 & 1.06 \\
 & DAD4TS & \textbf{\textcolor{red}{89.1}} & \textbf{\textcolor{red}{-34.5}} & \textbf{\textcolor{red}{0.454}} & \textbf{\textcolor{red}{3.29}} & \textbf{\textcolor{red}{0.376}} & \textbf{\textcolor{red}{2.94}} & \textbf{\textcolor{red}{0.540}} & \textbf{\textcolor{red}{3.98}} & 0.144 & 0.986 & 0.143 & 0.933 & 0.144 & 0.989 & 0.153 & 1.06 & \textcolor{blue}{0.142} & \textcolor{blue}{0.941} \\
Forest & baseline & 0.00 & 0.00 & 1.15 & 7.39 & 1.27 & 8.42 & \textbf{\textcolor{red}{1.08}} & \textcolor{blue}{7.10} & \textcolor{blue}{1.02} & \textcolor{blue}{6.12} & 1.10 & 6.41 & \textbf{\textcolor{red}{1.08}} & 6.31 & \textcolor{blue}{1.02} & \textbf{\textcolor{red}{6.18}} & \textbf{\textcolor{red}{1.03}} & \textbf{\textcolor{red}{6.29}} \\
 & TimeGAN & 37.5 & 1.62 & 1.16 & 7.17 & 1.21 & 7.81 & 1.15 & 7.54 & 1.04 & 6.24 & \textbf{\textcolor{red}{1.06}} & \textbf{\textcolor{red}{6.22}} & 1.09 & 6.33 & 1.08 & 6.59 & 1.05 & 6.73 \\
 & TimeVAE & 35.9 & 1.11 & 1.14 & 7.24 & 1.24 & 8.25 & 1.14 & 7.41 & 1.02 & 6.14 & 1.12 & 6.41 & 1.12 & 6.34 & 1.03 & 6.32 & 1.05 & \textcolor{blue}{6.33} \\
 & SMFG & - & - & - & - & - & - & - & - & - & - & - & - & - & - & - & - & - & - \\
 & Diffusion-TS & \textbf{\textcolor{red}{45.3}} & \textcolor{blue}{-1.27} & \textcolor{blue}{1.08} & \textcolor{blue}{6.98} & \textcolor{blue}{1.11} & \textcolor{blue}{7.36} & 1.13 & 7.49 & 1.02 & 6.22 & 1.10 & 6.46 & \textcolor{blue}{1.09} & \textbf{\textcolor{red}{6.18}} & \textbf{\textcolor{red}{1.02}} & 6.20 & \textcolor{blue}{1.05} & 6.66 \\
 & ReAugment & 39.1 & 0.790 & 1.21 & 7.53 & 1.15 & 7.71 & \textcolor{blue}{1.09} & \textbf{\textcolor{red}{7.03}} & 1.06 & 6.27 & \textcolor{blue}{1.09} & 6.34 & 1.12 & 6.35 & 1.04 & \textcolor{blue}{6.18} & 1.06 & 6.62 \\
 & DiffAT & 37.5 & 1.78 & 1.20 & 7.48 & 1.27 & 8.46 & 1.10 & 7.29 & \textbf{\textcolor{red}{1.00}} & \textbf{\textcolor{red}{6.04}} & 1.10 & 6.53 & 1.10 & 6.44 & 1.05 & 6.28 & 1.07 & 6.77 \\
 & AutoDA & 32.8 & 1.24 & 1.20 & 7.51 & 1.24 & 8.22 & 1.10 & 7.28 & 1.05 & 6.20 & 1.09 & \textcolor{blue}{6.26} & 1.11 & \textcolor{blue}{6.22} & 1.05 & 6.27 & 1.06 & 6.66 \\
 & DAD4TS & \textcolor{blue}{45.3} & \textbf{\textcolor{red}{-2.33}} & \textbf{\textcolor{red}{1.05}} & \textbf{\textcolor{red}{6.66}} & \textbf{\textcolor{red}{1.06}} & \textbf{\textcolor{red}{6.83}} & 1.10 & 7.32 & 1.03 & 6.24 & 1.10 & 6.52 & 1.09 & 6.40 & 1.04 & 6.21 & 1.05 & 6.46 \\
German & baseline & 0.00 & 0.00 & 1.20 & 8.42 & 1.44 & 10.2 & 1.34 & 9.46 & 1.34 & 9.77 & 1.47 & 10.8 & 1.37 & 9.90 & 1.48 & 11.3 & \textbf{\textcolor{red}{1.39}} & \textcolor{blue}{10.2} \\
 & TimeGAN & 46.9 & 2.74 & 1.19 & 8.56 & 1.45 & 10.00 & \textcolor{blue}{1.30} & \textbf{\textcolor{red}{8.98}} & 1.35 & \textcolor{blue}{9.56} & 1.47 & 10.7 & 1.35 & 9.73 & \textbf{\textcolor{red}{1.33}} & \textcolor{blue}{9.67} & 1.59 & 11.7 \\
 & TimeVAE & 53.1 & 1.20 & 1.18 & 8.36 & \textcolor{blue}{1.41} & 9.74 & 1.33 & 9.48 & 1.37 & 10.00 & 1.44 & 10.6 & \textcolor{blue}{1.35} & \textbf{\textcolor{red}{9.63}} & 1.49 & 10.2 & 1.52 & 11.6 \\
 & SMFG & 45.3 & 1.44 & \textcolor{blue}{1.14} & \textcolor{blue}{8.02} & 1.52 & 11.0 & 1.32 & 9.21 & 1.41 & 10.2 & 1.41 & \textcolor{blue}{10.4} & 1.46 & 10.8 & 1.40 & 10.2 & 1.43 & 10.5 \\
 & Diffusion-TS & 39.1 & 3.62 & 1.18 & 8.07 & 1.42 & 10.00 & \textbf{\textcolor{red}{1.28}} & 9.26 & 1.42 & 10.4 & 1.49 & 10.8 & 1.39 & 10.2 & 1.43 & 10.0 & 1.51 & 11.2 \\
 & ReAugment & 60.9 & 2.14 & 1.18 & 8.43 & 1.42 & 9.78 & 1.32 & \textcolor{blue}{9.11} & \textcolor{blue}{1.33} & \textbf{\textcolor{red}{9.54}} & 1.53 & 11.0 & 1.41 & 10.0 & \textcolor{blue}{1.33} & \textbf{\textcolor{red}{9.39}} & 1.51 & 11.3 \\
 & DiffAT & \textcolor{blue}{62.5} & \textcolor{blue}{-0.937} & 1.18 & 8.23 & 1.46 & 10.2 & 1.33 & 9.11 & 1.35 & 9.93 & \textcolor{blue}{1.40} & 10.5 & 1.38 & 10.0 & 1.38 & 10.0 & \textcolor{blue}{1.41} & 10.3 \\
 & AutoDA & 46.9 & 0.305 & 1.24 & 8.88 & 1.42 & \textcolor{blue}{9.71} & 1.34 & 9.72 & 1.39 & 10.1 & 1.51 & 11.3 & 1.38 & 9.89 & 1.42 & 10.2 & 1.47 & 11.1 \\
 & DAD4TS & \textbf{\textcolor{red}{71.9}} & \textbf{\textcolor{red}{-3.42}} & \textbf{\textcolor{red}{1.09}} & \textbf{\textcolor{red}{7.69}} & \textbf{\textcolor{red}{1.36}} & \textbf{\textcolor{red}{9.55}} & 1.34 & 9.88 & \textbf{\textcolor{red}{1.32}} & 9.73 & \textbf{\textcolor{red}{1.36}} & \textbf{\textcolor{red}{9.88}} & \textbf{\textcolor{red}{1.32}} & \textcolor{blue}{9.70} & 1.38 & 10.4 & 1.43 & \textbf{\textcolor{red}{10.2}} \\
ILI & baseline & 0.00 & 0.00 & 1.83 & 12.5 & 1.81 & 12.1 & 1.98 & 13.3 & 0.567 & 3.66 & 0.580 & 3.72 & 0.584 & 3.70 & 0.601 & 3.92 & 0.614 & 4.12 \\
 & TimeGAN & 37.5 & 0.865 & 1.83 & 12.5 & 1.84 & 12.5 & 2.06 & 14.0 & 0.560 & 3.57 & 0.568 & 3.66 & 0.597 & 3.81 & \textcolor{blue}{0.573} & \textbf{\textcolor{red}{3.61}} & 0.601 & 3.99 \\
 & TimeVAE & 46.9 & -0.568 & 1.70 & 11.7 & 1.83 & 12.0 & 1.91 & 12.5 & 0.588 & 3.83 & 0.592 & 3.84 & 0.629 & 4.07 & 0.587 & 3.74 & 0.590 & 3.85 \\
 & SMFG & - & - & - & - & - & - & - & - & - & - & - & - & - & - & - & - & - & - \\
 & Diffusion-TS & 50.0 & -0.833 & 1.76 & 12.3 & 1.72 & 11.8 & 2.03 & 13.5 & 0.570 & 3.66 & 0.594 & 3.88 & 0.607 & 3.96 & 0.582 & 3.76 & 0.589 & 4.01 \\
 & ReAugment & 28.1 & 1.25 & 1.80 & 12.2 & 1.81 & 12.1 & 1.99 & 12.7 & 0.584 & 3.80 & 0.597 & 3.91 & 0.607 & 3.91 & 0.599 & 3.79 & 0.591 & 3.91 \\
 & DiffAT & 43.8 & 1.46 & 1.90 & 13.2 & 1.86 & 12.3 & 2.02 & 13.6 & 0.579 & 3.77 & 0.594 & 3.79 & 0.615 & 3.91 & 0.597 & 3.91 & \textcolor{blue}{0.563} & \textcolor{blue}{3.73} \\
 & AutoDA & \textcolor{blue}{76.6} & \textcolor{blue}{-11.3} & \textcolor{blue}{1.44} & \textcolor{blue}{8.86} & \textcolor{blue}{1.27} & \textbf{\textcolor{red}{7.03}} & \textcolor{blue}{1.48} & \textcolor{blue}{9.84} & \textcolor{blue}{0.550} & \textbf{\textcolor{red}{3.45}} & \textcolor{blue}{0.544} & \textcolor{blue}{3.43} & \textcolor{blue}{0.554} & \textcolor{blue}{3.50} & \textbf{\textcolor{red}{0.573}} & \textcolor{blue}{3.64} & 0.576 & 3.76 \\
 & DAD4TS & \textbf{\textcolor{red}{96.9}} & \textbf{\textcolor{red}{-20.4}} & \textbf{\textcolor{red}{1.13}} & \textbf{\textcolor{red}{7.38}} & \textbf{\textcolor{red}{1.07}} & \textcolor{blue}{7.13} & \textbf{\textcolor{red}{1.12}} & \textbf{\textcolor{red}{7.41}} & \textbf{\textcolor{red}{0.539}} & \textcolor{blue}{3.45} & \textbf{\textcolor{red}{0.535}} & \textbf{\textcolor{red}{3.40}} & \textbf{\textcolor{red}{0.534}} & \textbf{\textcolor{red}{3.26}} & 0.575 & 3.66 & \textbf{\textcolor{red}{0.536}} & \textbf{\textcolor{red}{3.43}} \\
Inventories & baseline & 0.00 & 0.00 & 0.748 & 5.35 & 0.719 & 5.20 & 0.859 & 6.24 & 0.552 & 3.78 & \textbf{\textcolor{red}{0.548}} & \textbf{\textcolor{red}{3.64}} & \textbf{\textcolor{red}{0.567}} & \textcolor{blue}{3.83} & \textbf{\textcolor{red}{0.524}} & \textbf{\textcolor{red}{3.47}} & 0.561 & 3.82 \\
 & TimeGAN & 43.8 & 1.14 & 0.735 & 5.25 & 0.784 & 5.75 & 0.818 & 6.26 & \textbf{\textcolor{red}{0.548}} & \textbf{\textcolor{red}{3.73}} & 0.572 & 3.80 & 0.570 & 3.89 & 0.569 & 3.83 & \textbf{\textcolor{red}{0.527}} & \textcolor{blue}{3.60} \\
 & TimeVAE & 40.6 & -0.186 & 0.741 & 5.46 & 0.730 & 5.50 & 0.730 & 5.50 & 0.571 & 3.98 & 0.579 & 3.92 & 0.584 & 3.93 & 0.557 & 3.75 & 0.566 & 3.99 \\
 & SMFG & 35.9 & 3.62 & 0.755 & 5.45 & 0.756 & 5.21 & 0.897 & 6.28 & \textcolor{blue}{0.551} & \textcolor{blue}{3.74} & 0.584 & 3.91 & 0.584 & 3.98 & 0.532 & 3.68 & 0.577 & 4.00 \\
 & Diffusion-TS & 26.6 & 6.29 & 0.816 & 6.18 & 0.839 & 6.48 & 0.908 & 6.68 & 0.568 & 3.93 & 0.564 & 3.81 & 0.594 & 4.07 & 0.553 & 3.86 & 0.558 & 3.85 \\
 & ReAugment & 26.6 & 4.60 & 0.735 & \textcolor{blue}{5.22} & 0.746 & 5.38 & 0.865 & 6.50 & 0.592 & 4.11 & 0.585 & 3.97 & 0.598 & 4.09 & 0.573 & 3.92 & 0.575 & 4.01 \\
 & DiffAT & 45.3 & -1.48 & \textcolor{blue}{0.730} & 5.24 & 0.753 & 5.39 & 0.725 & 5.44 & 0.559 & 3.80 & 0.569 & 3.86 & 0.580 & 4.02 & 0.538 & 3.59 & \textcolor{blue}{0.529} & \textbf{\textcolor{red}{3.54}} \\
 & AutoDA & \textcolor{blue}{60.9} & \textcolor{blue}{-4.90} & 0.738 & 5.72 & \textcolor{blue}{0.637} & \textcolor{blue}{4.60} & \textcolor{blue}{0.687} & \textcolor{blue}{5.31} & 0.554 & 3.76 & \textcolor{blue}{0.558} & \textcolor{blue}{3.77} & \textcolor{blue}{0.568} & \textbf{\textcolor{red}{3.77}} & \textcolor{blue}{0.531} & \textcolor{blue}{3.50} & 0.550 & 3.71 \\
 & DAD4TS & \textbf{\textcolor{red}{62.5}} & \textbf{\textcolor{red}{-12.8}} & \textbf{\textcolor{red}{0.469}} & \textbf{\textcolor{red}{3.23}} & \textbf{\textcolor{red}{0.547}} & \textbf{\textcolor{red}{4.20}} & \textbf{\textcolor{red}{0.508}} & \textbf{\textcolor{red}{3.57}} & 0.552 & 3.77 & 0.566 & 3.81 & 0.586 & 3.97 & 0.551 & 3.78 & 0.560 & 3.92 \\
Consump. & baseline & 0.00 & 0.00 & 4.71 & 36.6 & 4.58 & 35.7 & 4.51 & 34.9 & 0.342 & 1.46 & 0.398 & \textcolor{blue}{1.95} & 0.394 & \textcolor{blue}{1.51} & 0.688 & 3.95 & 0.529 & 2.95 \\
 & TimeGAN & 37.5 & 15.2 & 5.03 & 40.5 & 5.99 & 47.7 & 4.74 & 38.1 & 0.379 & 1.55 & 0.463 & 2.44 & 0.397 & 1.74 & 0.736 & 3.75 & 0.552 & 3.50 \\
 & TimeVAE & 51.6 & 7.52 & 4.72 & 36.5 & \textcolor{blue}{4.51} & \textbf{\textcolor{red}{35.1}} & 4.37 & 34.0 & 0.475 & 2.02 & 0.462 & 2.41 & 0.435 & 1.70 & 0.709 & 4.02 & \textbf{\textcolor{red}{0.342}} & \textbf{\textcolor{red}{1.62}} \\
 & SMFG & 40.6 & 10.4 & 4.75 & 37.0 & 5.11 & 41.6 & \textcolor{blue}{4.36} & \textcolor{blue}{33.5} & \textcolor{blue}{0.328} & \textcolor{blue}{1.39} & 0.575 & 3.29 & 0.390 & 1.68 & 0.711 & 4.23 & 0.638 & 5.31 \\
 & Diffusion-TS & 46.9 & \textcolor{blue}{-1.62} & 5.03 & 39.2 & 4.82 & 38.0 & 4.86 & 38.2 & 0.336 & 1.42 & \textcolor{blue}{0.387} & 1.95 & \textcolor{blue}{0.344} & 1.55 & \textcolor{blue}{0.401} & \textcolor{blue}{1.76} & 0.480 & 2.81 \\
 & ReAugment & \textcolor{blue}{54.7} & 3.61 & \textcolor{blue}{4.64} & \textbf{\textcolor{red}{35.8}} & 4.56 & 35.3 & 4.47 & 34.5 & 0.453 & 1.94 & \textbf{\textcolor{red}{0.370}} & \textbf{\textcolor{red}{1.81}} & 0.411 & 1.68 & 0.644 & 3.16 & 0.503 & 2.90 \\
 & DiffAT & 42.2 & 2.09 & 4.80 & 37.5 & 4.59 & 35.8 & 4.48 & 34.5 & 0.429 & 1.77 & 0.406 & 2.09 & 0.348 & 1.54 & 0.569 & 3.06 & \textcolor{blue}{0.452} & \textcolor{blue}{2.18} \\
 & AutoDA & \textbf{\textcolor{red}{64.1}} & \textbf{\textcolor{red}{-5.42}} & \textbf{\textcolor{red}{4.32}} & \textcolor{blue}{36.1} & \textbf{\textcolor{red}{4.22}} & \textcolor{blue}{35.2} & \textbf{\textcolor{red}{3.88}} & \textbf{\textcolor{red}{31.3}} & \textbf{\textcolor{red}{0.312}} & \textbf{\textcolor{red}{1.31}} & 0.469 & 2.42 & 0.370 & 1.52 & \textbf{\textcolor{red}{0.391}} & \textbf{\textcolor{red}{1.68}} & 0.570 & 3.40 \\
 & DAD4TS & 43.8 & 2.82 & 5.09 & 40.8 & 5.71 & 43.2 & 5.41 & 42.4 & 0.344 & 1.40 & 0.435 & 2.22 & \textbf{\textcolor{red}{0.323}} & \textbf{\textcolor{red}{1.39}} & 0.429 & 2.26 & 0.554 & 3.11 \\
\hline
\end{tabular}
}
\end{table}


Table ~\ref{all_avg4pred_rmse} reports the comprehensive experimental results of DAD4TS. The best results are highlighted in red, and the second-best results are highlighted in blue with underlines. The \textbf{baseline} indicates the accuracy of the predictive model when trained using only real data. Imp. Rate (\%) indicates the percentage of cases in which the forecasting accuracy is improved across all settings, and Imp. Mean (\%) denotes the average improvement in forecasting accuracy.

\subsection{Model Analysis}
\paragraph{Distribution of generated data.}

\begin{figure}
    \centering
        \includegraphics[
      width=1.0\linewidth,
    ]{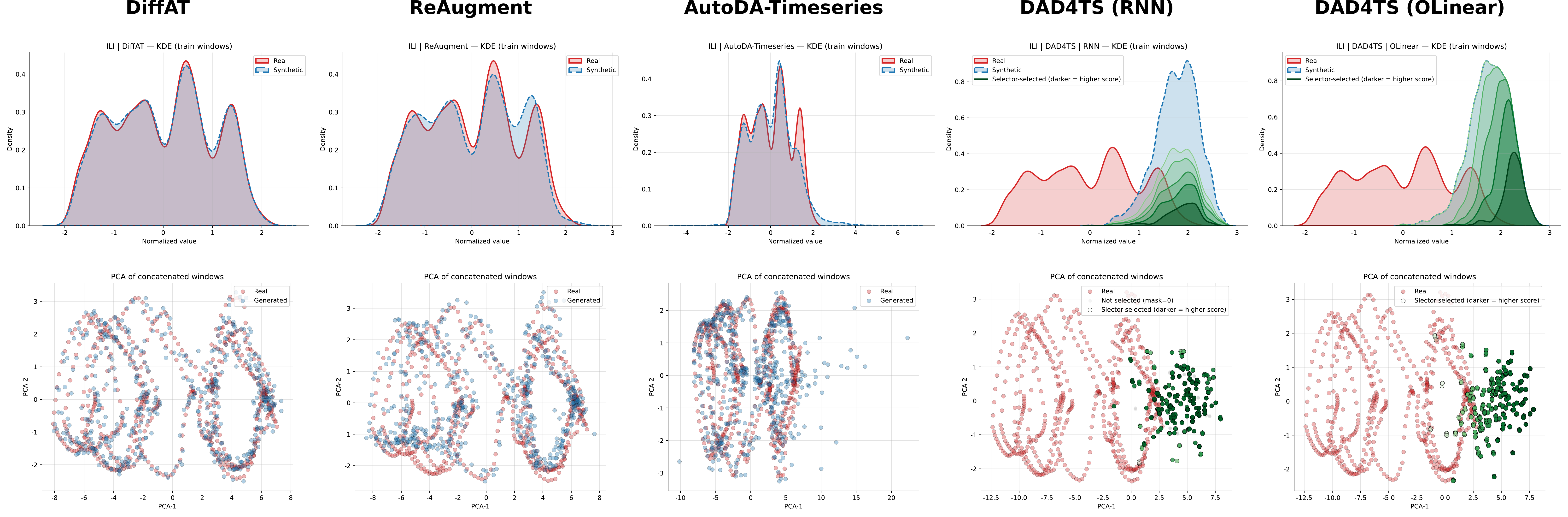}
    \caption{Distribution of generated data for each representative method. Red indicates real data, and blue indicates generated data. Green indicates generated data selected by the Selector as suitable for training. The darker the color, the higher the data is rated as valuable for learning. In this figure, the ILI dataset is used, and the input and forecast lengths are fixed at 12.}
    \label{compare_distribution}
\end{figure}

Figure~\ref{compare_distribution} shows the data distributions of the samples generated by three representative baseline methods and by the proposed method. From Figure~\ref{compare_distribution}, we can see that the baseline methods overall generate data whose distribution is similar to that of the real data, whereas the proposed method generates data whose distribution is slightly different from the real data. This is presumably because, in DAD4TS, training and sampling with the diffusion model are performed on the geometric space constructed by PCA, which leads to generation that emphasizes features with particularly large variance in the time-series data. We also observe that the distributions of the data generated by the diffusion model differ to some extent across TSF models, and that the samples selected by the Selector as more valuable training data, as well as their quantities, also vary. These observations suggest that the Selector assesses the generated data as valuable for training in a manner tailored to each TSF model. In addition, the results confirm that not all generated data are used and only a subset is utilized; thus, the Selector can subsample generated data that would act as noise during training while selecting and training on data suited to the TSF model, which is considered to have improved the prediction accuracy of the TSF model as a result.


\setlength{\tabcolsep}{1pt}
\renewcommand{\arraystretch}{0.90}
\begin{table}[t]
\footnotesize
\centering
\caption{Ablation study: Average RMSE and DTW over multiple forecast lengths with input length fixed to 12. Results are averaged over forecast lengths $\in\{3,6,9,12\}$ for all datasets, except ILI where $\in\{2,4,8,12\}$. R represents RMSE, and D represents DTW. DAD4TS Once indicates a setting in which data generation is performed only once.}
\label{abl_DAD4TS}
\resizebox{\linewidth}{!}{%
\begin{tabular}{@{} l l r r *{16}{c} @{}}
\hline
Dataset & Methods & \multicolumn{2}{c}{Imp.(\%)} & \multicolumn{2}{c}{RNN} & \multicolumn{2}{c}{LSTM} & \multicolumn{2}{c}{Trans.} & \multicolumn{2}{c}{iTrans.} & \multicolumn{2}{c}{Patch.} & \multicolumn{2}{c}{Times.} & \multicolumn{2}{c}{S2IP.} & \multicolumn{2}{c}{OLin.} \\
 &  & Rate $\uparrow$ & Mean $\downarrow$ & R $\downarrow$ & D $\downarrow$ & R $\downarrow$ & D $\downarrow$ & R $\downarrow$ & D $\downarrow$ & R $\downarrow$ & D $\downarrow$ & R $\downarrow$ & D $\downarrow$ & R $\downarrow$ & D $\downarrow$ & R $\downarrow$ & D $\downarrow$ & R $\downarrow$ & D $\downarrow$ \\
\hline
Employees & baseline & 0.00 & 0.00 & 1.32 & 10.1 & 1.38 & 10.5 & 1.30 & 9.67 & 0.153 & 1.12 & 0.144 & 0.953 & 0.192 & 1.49 & 0.184 & 1.31 & 0.201 & 1.47 \\
 & DAD4TS Once & \textbf{\textcolor{red}{92.2}} & \textcolor{blue}{-30.4} & \textcolor{blue}{0.532} & \textcolor{blue}{3.97} & \textcolor{blue}{0.643} & \textcolor{blue}{4.91} & \textcolor{blue}{0.633} & \textcolor{blue}{4.89} & \textbf{\textcolor{red}{0.141}} & \textcolor{blue}{1.00} & \textbf{\textcolor{red}{0.139}} & \textbf{\textcolor{red}{0.910}} & \textcolor{blue}{0.147} & \textcolor{blue}{1.01} & \textbf{\textcolor{red}{0.152}} & \textbf{\textcolor{red}{1.03}} & \textcolor{blue}{0.154} & \textcolor{blue}{1.04} \\
 & DAD4TS w/o Selector & 59.4 & -3.24 & 1.35 & 10.3 & 1.35 & 10.2 & 1.27 & 9.60 & 0.156 & 1.12 & 0.143 & 0.954 & 0.154 & 1.10 & 0.210 & 1.51 & 0.166 & 1.22 \\
 & DAD4TS & \textcolor{blue}{89.1} & \textbf{\textcolor{red}{-34.5}} & \textbf{\textcolor{red}{0.454}} & \textbf{\textcolor{red}{3.29}} & \textbf{\textcolor{red}{0.376}} & \textbf{\textcolor{red}{2.94}} & \textbf{\textcolor{red}{0.540}} & \textbf{\textcolor{red}{3.98}} & \textcolor{blue}{0.144} & \textbf{\textcolor{red}{0.986}} & \textcolor{blue}{0.143} & \textcolor{blue}{0.933} & \textbf{\textcolor{red}{0.144}} & \textbf{\textcolor{red}{0.989}} & \textcolor{blue}{0.153} & \textcolor{blue}{1.06} & \textbf{\textcolor{red}{0.142}} & \textbf{\textcolor{red}{0.941}} \\
Forest & baseline & 0.00 & 0.00 & 1.15 & 7.39 & 1.27 & 8.42 & \textcolor{blue}{1.08} & \textcolor{blue}{7.10} & \textcolor{blue}{1.02} & \textbf{\textcolor{red}{6.12}} & 1.10 & \textcolor{blue}{6.41} & \textcolor{blue}{1.08} & 6.31 & \textbf{\textcolor{red}{1.02}} & \textbf{\textcolor{red}{6.18}} & \textbf{\textcolor{red}{1.03}} & \textbf{\textcolor{red}{6.29}} \\
 & DAD4TS Once & \textcolor{blue}{43.8} & \textcolor{blue}{-2.23} & \textcolor{blue}{1.08} & \textbf{\textcolor{red}{6.65}} & \textbf{\textcolor{red}{1.05}} & \textbf{\textcolor{red}{6.73}} & \textbf{\textcolor{red}{1.07}} & \textbf{\textcolor{red}{7.01}} & 1.03 & 6.25 & \textcolor{blue}{1.10} & 6.46 & 1.09 & \textcolor{blue}{6.25} & 1.05 & 6.33 & 1.04 & 6.64 \\
 & DAD4TS w/o Selector & 39.1 & 0.234 & 1.15 & 7.38 & 1.27 & 8.46 & 1.09 & 7.14 & \textbf{\textcolor{red}{1.01}} & \textcolor{blue}{6.14} & \textbf{\textcolor{red}{1.05}} & \textbf{\textcolor{red}{6.29}} & \textbf{\textcolor{red}{1.08}} & \textbf{\textcolor{red}{6.15}} & 1.05 & 6.46 & \textcolor{blue}{1.04} & \textcolor{blue}{6.43} \\
 & DAD4TS & \textbf{\textcolor{red}{45.3}} & \textbf{\textcolor{red}{-2.33}} & \textbf{\textcolor{red}{1.05}} & \textcolor{blue}{6.66} & \textcolor{blue}{1.06} & \textcolor{blue}{6.83} & 1.10 & 7.32 & 1.03 & 6.24 & 1.10 & 6.52 & 1.09 & 6.40 & \textcolor{blue}{1.04} & \textcolor{blue}{6.21} & 1.05 & 6.46 \\
German & baseline & 0.00 & 0.00 & 1.20 & 8.42 & 1.44 & 10.2 & 1.34 & \textcolor{blue}{9.46} & \textcolor{blue}{1.34} & \textcolor{blue}{9.77} & 1.47 & 10.8 & 1.37 & 9.90 & 1.48 & 11.3 & \textbf{\textcolor{red}{1.39}} & 10.2 \\
 & DAD4TS Once & \textcolor{blue}{62.5} & \textcolor{blue}{-1.32} & \textbf{\textcolor{red}{1.05}} & \textbf{\textcolor{red}{7.18}} & \textcolor{blue}{1.40} & \textcolor{blue}{10.1} & \textbf{\textcolor{red}{1.30}} & \textbf{\textcolor{red}{9.46}} & 1.38 & 10.1 & 1.44 & 10.5 & \textcolor{blue}{1.37} & 10.0 & \textcolor{blue}{1.35} & \textcolor{blue}{9.86} & \textcolor{blue}{1.40} & \textbf{\textcolor{red}{10.1}} \\
 & DAD4TS w/o Selector & 40.6 & 3.92 & 1.22 & 8.23 & 1.50 & 10.3 & 1.43 & 10.7 & 1.40 & 10.2 & \textcolor{blue}{1.43} & \textcolor{blue}{10.4} & 1.37 & \textcolor{blue}{9.86} & \textbf{\textcolor{red}{1.33}} & \textbf{\textcolor{red}{9.17}} & 1.54 & 11.7 \\
 & DAD4TS & \textbf{\textcolor{red}{71.9}} & \textbf{\textcolor{red}{-3.42}} & \textcolor{blue}{1.09} & \textcolor{blue}{7.69} & \textbf{\textcolor{red}{1.36}} & \textbf{\textcolor{red}{9.55}} & \textcolor{blue}{1.34} & 9.88 & \textbf{\textcolor{red}{1.32}} & \textbf{\textcolor{red}{9.73}} & \textbf{\textcolor{red}{1.36}} & \textbf{\textcolor{red}{9.88}} & \textbf{\textcolor{red}{1.32}} & \textbf{\textcolor{red}{9.70}} & 1.38 & 10.4 & 1.43 & \textcolor{blue}{10.2} \\
ILI & baseline & 0.00 & 0.00 & 1.83 & 12.5 & 1.81 & 12.1 & 1.98 & 13.3 & 0.567 & 3.66 & 0.580 & 3.72 & 0.584 & 3.70 & 0.601 & 3.92 & 0.614 & 4.12 \\
 & DAD4TS Once & \textcolor{blue}{81.2} & \textcolor{blue}{-19.0} & \textbf{\textcolor{red}{1.10}} & \textcolor{blue}{7.41} & \textcolor{blue}{1.07} & \textbf{\textcolor{red}{6.75}} & \textcolor{blue}{1.19} & \textcolor{blue}{7.51} & 0.556 & \textcolor{blue}{3.51} & \textbf{\textcolor{red}{0.531}} & \textbf{\textcolor{red}{3.34}} & 0.560 & \textcolor{blue}{3.48} & \textbf{\textcolor{red}{0.532}} & \textbf{\textcolor{red}{3.37}} & 0.566 & 3.69 \\
 & DAD4TS w/o Selector & 60.9 & -2.39 & 1.73 & 11.9 & 1.86 & 12.9 & 2.00 & 13.3 & \textcolor{blue}{0.551} & 3.54 & 0.546 & 3.44 & \textcolor{blue}{0.552} & 3.54 & 0.583 & 3.78 & \textcolor{blue}{0.556} & \textcolor{blue}{3.55} \\
 & DAD4TS & \textbf{\textcolor{red}{96.9}} & \textbf{\textcolor{red}{-20.4}} & \textcolor{blue}{1.13} & \textbf{\textcolor{red}{7.38}} & \textbf{\textcolor{red}{1.07}} & \textcolor{blue}{7.13} & \textbf{\textcolor{red}{1.12}} & \textbf{\textcolor{red}{7.41}} & \textbf{\textcolor{red}{0.539}} & \textbf{\textcolor{red}{3.45}} & \textcolor{blue}{0.535} & \textcolor{blue}{3.40} & \textbf{\textcolor{red}{0.534}} & \textbf{\textcolor{red}{3.26}} & \textcolor{blue}{0.575} & \textcolor{blue}{3.66} & \textbf{\textcolor{red}{0.536}} & \textbf{\textcolor{red}{3.43}} \\
Inventories & baseline & 0.00 & 0.00 & 0.748 & 5.35 & 0.719 & 5.20 & 0.859 & 6.24 & \textcolor{blue}{0.552} & 3.78 & \textbf{\textcolor{red}{0.548}} & \textbf{\textcolor{red}{3.64}} & \textcolor{blue}{0.567} & \textcolor{blue}{3.83} & \textbf{\textcolor{red}{0.524}} & \textbf{\textcolor{red}{3.47}} & 0.561 & \textcolor{blue}{3.82} \\
 & DAD4TS Once & \textcolor{blue}{59.4} & \textcolor{blue}{-11.6} & \textbf{\textcolor{red}{0.461}} & \textbf{\textcolor{red}{3.16}} & \textbf{\textcolor{red}{0.514}} & \textbf{\textcolor{red}{3.76}} & \textcolor{blue}{0.535} & \textcolor{blue}{3.77} & 0.554 & 3.80 & 0.583 & 3.92 & 0.587 & 4.06 & 0.582 & 3.93 & \textcolor{blue}{0.548} & 3.91 \\
 & DAD4TS w/o Selector & 50.0 & 3.05 & 0.819 & 5.91 & 0.935 & 7.02 & 0.711 & 5.25 & \textbf{\textcolor{red}{0.538}} & \textbf{\textcolor{red}{3.64}} & \textcolor{blue}{0.556} & \textcolor{blue}{3.75} & \textbf{\textcolor{red}{0.558}} & \textbf{\textcolor{red}{3.75}} & \textcolor{blue}{0.538} & \textcolor{blue}{3.58} & \textbf{\textcolor{red}{0.541}} & \textbf{\textcolor{red}{3.70}} \\
 & DAD4TS & \textbf{\textcolor{red}{62.5}} & \textbf{\textcolor{red}{-12.8}} & \textcolor{blue}{0.469} & \textcolor{blue}{3.23} & \textcolor{blue}{0.547} & \textcolor{blue}{4.20} & \textbf{\textcolor{red}{0.508}} & \textbf{\textcolor{red}{3.57}} & 0.552 & \textcolor{blue}{3.77} & 0.566 & 3.81 & 0.586 & 3.97 & 0.551 & 3.78 & 0.560 & 3.92 \\
Consump. & baseline & 0.00 & 0.00 & \textbf{\textcolor{red}{4.71}} & \textbf{\textcolor{red}{36.6}} & \textbf{\textcolor{red}{4.58}} & \textbf{\textcolor{red}{35.7}} & \textcolor{blue}{4.51} & \textcolor{blue}{34.9} & \textbf{\textcolor{red}{0.342}} & \textcolor{blue}{1.46} & 0.398 & 1.95 & 0.394 & \textcolor{blue}{1.51} & 0.688 & 3.95 & 0.529 & \textcolor{blue}{2.95} \\
 & DAD4TS Once & 37.5 & \textcolor{blue}{6.54} & 5.11 & 41.2 & 5.71 & 43.3 & 5.16 & 41.1 & 0.402 & 1.66 & \textbf{\textcolor{red}{0.344}} & \textbf{\textcolor{red}{1.69}} & \textcolor{blue}{0.326} & 1.54 & \textcolor{blue}{0.575} & 2.91 & \textbf{\textcolor{red}{0.406}} & \textbf{\textcolor{red}{1.94}} \\
 & DAD4TS w/o Selector & \textcolor{blue}{40.6} & 8.55 & 5.53 & 42.9 & 5.99 & 47.2 & \textbf{\textcolor{red}{4.40}} & \textbf{\textcolor{red}{33.9}} & 0.381 & 1.70 & \textcolor{blue}{0.351} & \textcolor{blue}{1.71} & 0.385 & 1.90 & 0.626 & \textcolor{blue}{2.90} & \textcolor{blue}{0.525} & 3.06 \\
 & DAD4TS & \textbf{\textcolor{red}{43.8}} & \textbf{\textcolor{red}{2.82}} & \textcolor{blue}{5.09} & \textcolor{blue}{40.8} & \textcolor{blue}{5.71} & \textcolor{blue}{43.2} & 5.41 & 42.4 & \textcolor{blue}{0.344} & \textbf{\textcolor{red}{1.40}} & 0.435 & 2.22 & \textbf{\textcolor{red}{0.323}} & \textbf{\textcolor{red}{1.39}} & \textbf{\textcolor{red}{0.429}} & \textbf{\textcolor{red}{2.26}} & 0.554 & 3.11 \\
\hline
\end{tabular}
}
\end{table}

\paragraph{Ablation Study.}

Compared to many conventional methods, DAD4TS sequentially generates synthetic data in tandem with the training of the TSF model. It also utilizes a Selector to determine whether the generated data is valuable for training the TSF model. To investigate how each of these components affects the performance of DAD4TS, we compared DAD4TS with versions where specific conditions were restricted. Specifically, we prepared a version where data generation was performed only once and a version where the Selector was removed from DAD4TS. The results are shown in Table~\ref{abl_DAD4TS}. Table~\ref{abl_DAD4TS} confirms that the Selector is necessary for the sequential generation of synthetic data in sync with the training of the time-series forecasting model. This demonstrates that the Selector functions effectively even when the amount of trainable data is small. Furthermore, it confirms that in environments where the Selector is available, sequentially generating synthetic data in sync with the forecasting model—rather than performing a one-time static generation—enables the improvement of accuracy in downstream tasks.

\section{Conclusion}

In this paper, we propose DAD4TS, a data augmentation framework for time-series forecasting in the small-scale data regime. This framework employs a ‘joint training strategy’ that simultaneously trains and updates, in an online manner, ‘time-series forecasting models’, a ‘time-series data generation model’ implemented based on diffusion models, and a Selector that continuously selects useful generated samples based on prediction-oriented rewards. We have demonstrated that, compared to existing data augmentation methods, this approach significantly improves the accuracy of time-series forecasting even under conditions of small-scale data. This framework imposes no restrictions on the time-series forecasting models. In future work, to address the limitations of the proposed method’s applicability—restricted to univariate time-series forecasting tasks—we are going to aim to extend it to downstream tasks beyond time-series forecasting as well as to multivariate data settings.

\medskip

{
\small

\bibliographystyle{unsrtnat}
\bibliography{main}

@INPROCEEDINGS{Liu_elec,
  author={Liu, Hengbo and Ma, Ziqing and Yang, Linxiao and Zhou, Tian and Xia, Rui and Wang, Yi and Wen, Qingsong and Sun, Liang},
  booktitle={ICASSP 2023 - 2023 IEEE International Conference on Acoustics, Speech and Signal Processing (ICASSP)}, 
  title={SADI: A Self-Adaptive Decomposed Interpretable Framework for Electric Load Forecasting Under Extreme Events}, 
  year={2023},
  volume={},
  number={},
  pages={1-5},
  doi={10.1109/ICASSP49357.2023.10096002}}

@inproceedings{
li2018diffusion,
title={Diffusion Convolutional Recurrent Neural Network: Data-Driven Traffic Forecasting},
author={Yaguang Li and Rose Yu and Cyrus Shahabi and Yan Liu},
booktitle={International Conference on Learning Representations},
year={2018},
url={https://openreview.net/forum?id=SJiHXGWAZ},
}

@article{dimri2020time,
  title={Time series analysis of climate variables using seasonal ARIMA approach},
  author={Dimri, Tripti and Ahmad, Shamshad and Sharif, Mohammad},
  journal={Journal of Earth System Science},
  volume={129},
  number={1},
  pages={149},
  year={2020},
  publisher={Springer}
}

@article{WANG2021504,
title = {Deep time series models for scarce data},
journal = {Neurocomputing},
volume = {456},
pages = {504-518},
year = {2021},
issn = {0925-2312},
doi = {https://doi.org/10.1016/j.neucom.2020.12.132},
url = {https://www.sciencedirect.com/science/article/pii/S0925231221001922},
author = {Qiyao Wang and Ahmed Farahat and Chetan Gupta and Shuai Zheng}
}

@inproceedings{
goktas2025tempusbench,
title={TempusBench: An Evaluation Framework for Time-Series Forecasting},
author={Denizalp Goktas and Amy Greenwald and Gerardo Riano-Briceno and Alexandra Magnusson and Alif Abdullah and Beatriz de Lucio},
booktitle={Recent Advances in Time Series Foundation Models Have We Reached the 'BERT Moment'?},
year={2025},
url={https://openreview.net/forum?id=3fMa060Ag5}
}

@inbook{Jinsung_timegan,
author = {Yoon, Jinsung and Jarrett, Daniel and van der Schaar, Mihaela},
title = {Time-series generative adversarial networks},
year = {2019},
publisher = {Curran Associates Inc.},
address = {Red Hook, NY, USA},
booktitle = {Proceedings of the 33rd International Conference on Neural Information Processing Systems},
articleno = {494},
numpages = {11}
}

@inproceedings{donahue2019wavegan,
  title={Adversarial Audio Synthesis},
  author={Donahue, Chris and McAuley, Julian and Puckette, Miller},
  booktitle={ICLR},
  year={2019}
}

@article{xu2020cot,
  title={Cot-gan: Generating sequential data via causal optimal transport},
  author={Xu, Tianlin and Wenliang, Li Kevin and Munn, Michael and Acciaio, Beatrice},
  journal={Advances in neural information processing systems},
  volume={33},
  pages={8798--8809},
  year={2020}
}

@misc{
desai2022timevae,
title={Time{VAE}: A Variational Auto-Encoder for Multivariate Time Series Generation},
author={Abhyuday Desai and Cynthia Freeman and Zuhui Wang and Ian Beaver},
year={2022},
url={https://openreview.net/forum?id=VDdDvnwFoyM}
}

@INPROCEEDINGS{Masato8852250,
  author={Ishii, Masato and Sato, Atsushi},
  booktitle={2019 International Joint Conference on Neural Networks (IJCNN)}, 
  title={Training Deep Neural Networks with Adversarially Augmented Features for Small-scale Training Datasets}, 
  year={2019},
  volume={},
  number={},
  pages={1-8},
  doi={10.1109/IJCNN.2019.8852250}}

@article{chawla2002smote,
  title={SMOTE: synthetic minority over-sampling technique},
  author={Chawla, Nitesh V. and Bowyer, Kevin W. and Hall, Lawrence O. and Kegelmeyer, W. Philip},
  journal={Journal of artificial intelligence research},
  volume={16},
  pages={321--357},
  year={2002}
}

@INPROCEEDINGS{Masahiro10973545,
  author={Suzuki, Masahiro and Kodaka, Megumi and Fukazawa, Yusuke},
  booktitle={2024 IEEE/WIC International Conference on Web Intelligence and Intelligent Agent Technology (WI-IAT)}, 
  title={Synthetic Mobility Feature Generation for Mental Health Prediction using Diffusion Models}, 
  year={2024},
  volume={},
  number={},
  pages={102-109},
  doi={10.1109/WI-IAT62293.2024.00022}}

@inproceedings{
gonen2025time,
title={Time Series Generation Under Data Scarcity: A Unified Generative Modeling Approach},
author={Tal Gonen and Itai Pemper and Ilan Naiman and Nimrod Berman and Omri Azencot},
booktitle={The Thirty-ninth Annual Conference on Neural Information Processing Systems},
year={2025},
url={https://openreview.net/forum?id=p324ryBKTc}
}

@inproceedings{yuan2024diffusionts,
  title={Diffusion-{TS}: Interpretable Diffusion for General Time Series Generation},
  author={Xinyu Yuan and Yan Qiao},
  booktitle={The Twelfth International Conference on Learning Representations},
  year={2024},
  url={https://openreview.net/forum?id=4h1apFjO99}
}

@inproceedings{dvrl,
author = {Yoon, Jinsung and Arik, Sercan O. and Pfister, Tomas},
title = {Data valuation using reinforcement learning},
year = {2020},
publisher = {JMLR.org},
booktitle = {Proceedings of the 37th International Conference on Machine Learning},
articleno = {1005},
numpages = {10},
series = {ICML'20}
}

@inproceedings{
liu2022flow,
title={Flow Straight and Fast: Learning to Generate and Transfer Data with Rectified Flow},
author={Xingchao Liu and Chengyue Gong and qiang liu},
booktitle={The Eleventh International Conference on Learning Representations },
year={2023},
url={https://openreview.net/forum?id=XVjTT1nw5z}
}

@article{kingma2013auto,
  title={Auto-encoding variational bayes},
  author={Kingma, Diederik P and Welling, Max},
  journal={arXiv preprint arXiv:1312.6114},
  year={2013}
}

@inproceedings{rombach2022high,
  title={High-resolution image synthesis with latent diffusion models},
  author={Rombach, Robin and Blattmann, Andreas and Lorenz, Dominik and Esser, Patrick and Ommer, Bj{\"o}rn},
  booktitle={Proceedings of the IEEE/CVF conference on computer vision and pattern recognition},
  pages={10684--10695},
  year={2022}
}

@inproceedings{
chen2025dove,
title={{DOVE}: Efficient One-Step Diffusion Model for Real-World Video Super-Resolution},
author={Zheng Chen and Zichen Zou and Kewei Zhang and Xiongfei Su and Xin Yuan and Yong Guo and Yulun Zhang},
booktitle={The Thirty-ninth Annual Conference on Neural Information Processing Systems},
year={2025},
url={https://openreview.net/forum?id=DkJImu7t3A}
}

@inproceedings{
rahimi2025auggen,
title={AugGen: Synthetic Augmentation using Diffusion Models Can Improve Recognition},
author={Parsa Rahimi and Damien Teney and S{\'e}bastien Marcel},
booktitle={The Thirty-ninth Annual Conference on Neural Information Processing Systems},
year={2025},
url={https://openreview.net/forum?id=LuKlBH8DAT}
}

@article{MACKIEWICZ1993303,
title = {Principal components analysis (PCA)},
journal = {Computers \& Geosciences},
volume = {19},
number = {3},
pages = {303-342},
year = {1993},
issn = {0098-3004},
doi = {https://doi.org/10.1016/0098-3004(93)90090-R},
url = {https://www.sciencedirect.com/science/article/pii/009830049390090R},
author = {Andrzej Maćkiewicz and Waldemar Ratajczak}
}

@article{ansari2024chronos,
    title={Chronos: Learning the Language of Time Series},
    author={Ansari, Abdul Fatir and Stella, Lorenzo and Turkmen, Caner and Zhang, Xiyuan and Mercado, Pedro and Shen, Huibin and Shchur, Oleksandr and Rangapuram, Syama Syndar and Pineda Arango, Sebastian and Kapoor, Shubham and Zschiegner, Jasper and Maddix, Danielle C. and Mahoney, Michael W. and Torkkola, Kari and Gordon Wilson, Andrew and Bohlke-Schneider, Michael and Wang, Yuyang},
    journal={Transactions on Machine Learning Research},
    issn={2835-8856},
    year={2024},
    url={https://openreview.net/forum?id=gerNCVqqtR}
}

@inproceedings{
auer2025tirex,
title={TiRex: Zero-Shot Forecasting Across Long and Short Horizons with Enhanced In-Context Learning},
author={Andreas Auer and Patrick Podest and Daniel Klotz and Sebastian B{\"o}ck and G{\"u}nter Klambauer and Sepp Hochreiter},
booktitle={The Thirty-ninth Annual Conference on Neural Information Processing Systems},
year={2025},
url={https://openreview.net/forum?id=v7UqniC9pF}
}

@article{shu2024data,
  title={Data augmentation for seizure prediction with generative diffusion model},
  author={Shu, Kai and Wu, Le and Zhao, Yuchang and Liu, Aiping and Qian, Ruobing and Chen, Xun},
  journal={IEEE Transactions on Cognitive and Developmental Systems},
  volume={17},
  number={3},
  pages={577--591},
  year={2024},
  publisher={IEEE}
}

@article{zhang2025time,
  title={A time-series data augmentation model through diffusion and transformer integration},
  author={Zhang, Yuren and Pu, Zhongnan and Jing, Lei},
  journal={arXiv preprint arXiv:2505.03790},
  year={2025}
}

@article{ho2020denoising,
  title={Denoising diffusion probabilistic models},
  author={Ho, Jonathan and Jain, Ajay and Abbeel, Pieter},
  journal={Advances in neural information processing systems},
  volume={33},
  pages={6840--6851},
  year={2020}
}

@article{jiang2023opendataval,
  title={Opendataval: a unified benchmark for data valuation},
  author={Jiang, Kevin and Liang, Weixin and Zou, James Y and Kwon, Yongchan},
  journal={Advances in Neural Information Processing Systems},
  volume={36},
  pages={28624--28647},
  year={2023}
}

@inproceedings{
just2023lava,
title={{LAVA}: Data Valuation without Pre-Specified Learning Algorithms},
author={Hoang Anh Just and Feiyang Kang and Tianhao Wang and Yi Zeng and Myeongseob Ko and Ming Jin and Ruoxi Jia},
booktitle={The Eleventh International Conference on Learning Representations },
year={2023},
url={https://openreview.net/forum?id=JJuP86nBl4q}
}

@inproceedings{Vitaly2020,
author = {Feldman, Vitaly and Zhang, Chiyuan},
title = {What neural networks memorize and why: discovering the long tail via influence estimation},
year = {2020},
isbn = {9781713829546},
publisher = {Curran Associates Inc.},
address = {Red Hook, NY, USA},
booktitle = {Proceedings of the 34th International Conference on Neural Information Processing Systems},
articleno = {242},
numpages = {11},
location = {Vancouver, BC, Canada},
series = {NIPS '20}
}

@article{Kaveen2025,
author = {Hiniduma, Kaveen and Byna, Suren and Bez, Jean Luca},
title = {Data Readiness for AI: A 360-Degree Survey},
year = {2025},
issue_date = {September 2025},
publisher = {Association for Computing Machinery},
address = {New York, NY, USA},
volume = {57},
number = {9},
issn = {0360-0300},
url = {https://doi.org/10.1145/3722214},
doi = {10.1145/3722214},
journal = {ACM Comput. Surv.},
month = apr,
articleno = {219},
numpages = {39}
}

@article{kong2020diffwave,
  title={Diffwave: A versatile diffusion model for audio synthesis},
  author={Kong, Zhifeng and Ping, Wei and Huang, Jiaji and Zhao, Kexin and Catanzaro, Bryan},
  journal={arXiv preprint arXiv:2009.09761},
  year={2020}
}

@inproceedings{
ho2021classifierfree,
title={Classifier-Free Diffusion Guidance},
author={Jonathan Ho and Tim Salimans},
booktitle={NeurIPS 2021 Workshop on Deep Generative Models and Downstream Applications},
year={2021},
url={https://openreview.net/forum?id=qw8AKxfYbI}
}

@inproceedings{
deng2025oats,
title={{OATS}: Online Data Augmentation for Time Series Foundation Models},
author={Junwei Deng and Chang Xu and Jiaqi W. Ma and Jiang Bian},
booktitle={Recent Advances in Time Series Foundation Models Have We Reached the 'BERT Moment'?},
year={2025},
url={https://openreview.net/forum?id=kxdRkZqJLp}
}

@misc{yuan2025reaugmentmodelzooguidedrl,
      title={ReAugment: Model Zoo-Guided RL for Few-Shot Time Series Augmentation and Forecasting}, 
      author={Haochen Yuan and Yutong Wang and Yihong Chen and Yunbo Wang and Xiaokang Yang},
      year={2025},
      eprint={2409.06282},
      archivePrefix={arXiv},
      primaryClass={cs.LG},
      url={https://arxiv.org/abs/2409.06282}, 
}

@article{YU2025112091,
title = {DiffAT: Effective data augmentation with diffusion models for time series forecasting},
journal = {Engineering Applications of Artificial Intelligence},
volume = {161},
pages = {112091},
year = {2025},
issn = {0952-1976},
doi = {https://doi.org/10.1016/j.engappai.2025.112091},
url = {https://www.sciencedirect.com/science/article/pii/S0952197625020998},
author = {Yang Yu and Ruizhe Ma and Wenbo Gu and Zongmin Ma}
}

@article{TAN2026152,
title = {Data augmentation in time series forecasting through inverted framework},
journal = {Pattern Recognition Letters},
volume = {201},
pages = {152-159},
year = {2026},
issn = {0167-8655},
doi = {https://doi.org/10.1016/j.patrec.2026.01.019},
url = {https://www.sciencedirect.com/science/article/pii/S0167865526000279},
author = {Hongming Tan and Ting Chen and Ruochong Jin and Wai Kin Victor Chan}
}

@inproceedings{TarDiff,
author = {Deng, Bowen and Xu, Chang and Li, Hao and Huang, Yu-hao and Hou, Min and Bian, Jiang},
title = {TarDiff: Target-Oriented Diffusion Guidance for Synthetic Electronic Health Record Time Series Generation},
year = {2025},
isbn = {9798400714542},
publisher = {Association for Computing Machinery},
address = {New York, NY, USA},
url = {https://doi.org/10.1145/3711896.3737147},
doi = {10.1145/3711896.3737147},
booktitle = {Proceedings of the 31st ACM SIGKDD Conference on Knowledge Discovery and Data Mining V.2},
pages = {474–485},
numpages = {12},
location = {Toronto ON, Canada},
series = {KDD '25}
}

@inproceedings{
dou2026autodatimeseries,
title={Auto{DA}-Timeseries: Automated Data Augmentation for Time Series},
author={Zijun Dou and Zhenhe Yao and Zhe Xie and Xidao Wen and Tong Xiao and Dan Pei},
booktitle={The Fourteenth International Conference on Learning Representations},
year={2026},
url={https://openreview.net/forum?id=vTLmHAkoIW}
}

@article{hochreiter1997long,
  title={Long short-term memory},
  author={Hochreiter, Sepp and Schmidhuber, J{\"u}rgen},
  journal={Neural computation},
  volume={9},
  number={8},
  pages={1735--1780},
  year={1997},
  publisher={MIT press}
}

@article{vaswani2017attention,
  title={Attention is all you need},
  author={Vaswani, Ashish and Shazeer, Noam and Parmar, Niki and Uszkoreit, Jakob and Jones, Llion and Gomez, Aidan N and Kaiser, {\L}ukasz and Polosukhin, Illia},
  journal={Advances in neural information processing systems},
  volume={30},
  year={2017}
}

@article{liu2023itransformer,
  title={itransformer: Inverted transformers are effective for time series forecasting},
  author={Liu, Yong and Hu, Tengge and Zhang, Haoran and Wu, Haixu and Wang, Shiyu and Ma, Lintao and Long, Mingsheng},
  journal={arXiv preprint arXiv:2310.06625},
  year={2023}
}

@article{nie2022time,
  title={A time series is worth 64 words: Long-term forecasting with transformers},
  author={Nie, Yuqi and Nguyen, Nam H and Sinthong, Phanwadee and Kalagnanam, Jayant},
  journal={arXiv preprint arXiv:2211.14730},
  year={2022}
}

@article{wu2022timesnet,
  title={Timesnet: Temporal 2d-variation modeling for general time series analysis},
  author={Wu, Haixu and Hu, Tengge and Liu, Yong and Zhou, Hang and Wang, Jianmin and Long, Mingsheng},
  journal={arXiv preprint arXiv:2210.02186},
  year={2022}
}

@inproceedings{pan2024s,
  title={{S$^2$IP-LLM}: Semantic space informed prompt learning with LLM for time series forecasting},
  author={Pan, Zijie and Jiang, Yushan and Garg, Sahil and Schneider, Anderson and Nevmyvaka, Yuriy and Song, Dongjin},
  booktitle={Forty-first International Conference on Machine Learning},
  year={2024}
}

@inproceedings{
yue2025olinear,
title={{OL}inear: A Linear Model for Time Series Forecasting in Orthogonally Transformed Domain},
author={Wenzhen Yue and Yong Liu and Hao Wang and Haoxuan Li and Xianghua Ying and Ruohao Guo and Bowei Xing and Ji Shi},
booktitle={The Thirty-ninth Annual Conference on Neural Information Processing Systems},
year={2025},
url={https://openreview.net/forum?id=DAyKP1tvwI}
}

@misc{Inventories,
  author = {{U.S. Census Bureau. Manufacturers: Inventories to Sales Ratio [MNFCTRIRSA]}},
  year   = {2026},
  url    = {https://fred.stlouisfed.org/series/MNFCTRIRSA},
  note   = {Retrieved February 20, 2026}
}

@misc{German,
  author = {{Bank for International Settlements. Manufacturers: Real Residential Property Prices for Germany [QDER628BIS]}},
  year   = {2026},
  url    = {https://fred.stlouisfed.org/series/QDER628BIS},
  note   = {Retrieved February 20, 2026}
}

@misc{pcom,
  author = {{U.S. Bureau of Economic Analysis. Personal Consumption Expenditures: Chain-type Price Index [DPCERG3A086NBEA]}},
  year   = {2026},
  url    = {https://fred.stlouisfed.org/series/DPCERG3A086NBEA},
  note   = {Retrieved February 20, 2026}
}

@misc{Employees,
  author = {{U.S. Bureau of Labor Statistics. All Employees, Health Care [CES6562000101]}},
  year   = {2026},
  url    = {https://fred.stlouisfed.org/series/CES6562000101},
  note   = {Retrieved February 20, 2026}
}

@misc{forest_fires_162,
  author       = {Cortez, Paulo and Morais, Anbal},
  title        = {{Forest Fires}},
  year         = {2007},
  howpublished = {UCI Machine Learning Repository},
  note         = {{DOI}: https://doi.org/10.24432/C5D88D}
}

@article{wen2020time,
  title={Time series data augmentation for deep learning: A survey},
  author={Wen, Qingsong and Sun, Liang and Yang, Fan and Song, Xiaomin and Gao, Jingkun and Wang, Xue and Xu, Huan},
  journal={arXiv preprint arXiv:2002.12478},
  year={2020}
}

}


\appendix

\section{Dataset}
\label{dataset}

Wang et al. \citep{WANG2021504} implicitly defined a dataset of approximately $100$ records as small-scale, and Gonen et al. \citep{gonen2025time}, who proposed a data generation method for small-scale time-series data, utilised datasets of $1,000$ records or fewer for time-series prediction in few-shot tasks; consequently, we prepared a dataset of approximately $100$–$1,000$ records. Furthermore, regarding the selection criteria for the data used, as recent models might have employed methods tuned specifically to popular benchmark datasets, we selected the widely used \textbf{ILI} dataset \citep{yue2025olinear} alongside the relatively newly proposed benchmark dataset \citep{goktas2025tempusbench}, specifically the univariate dataset \textbf{Inventories to Sales Ratio} \citep {Inventories}, \textbf{German House Prices} \citep{German}, \textbf{Personal Consumption Expenditures} \citep{pcom}, \textbf{Employees Health Care} \citep{Employees}, and \textbf{Forest Fires} \citep{forest_fires_162} from the relatively recently proposed benchmark dataset \citep{goktas2025tempusbench}.

\section{Evaluation Metrics}
\label{metrics}

The evaluation metrics used were Root Mean Squared Error (RMSE), to measure the accuracy of time-series forecasting, and Dynamic Time Warping (DTW), to measure the structural similarity of time-series data, which cannot be captured by RMSE alone. By employing these two metrics, we evaluate the extent to which the time-series forecasting model is able to make accurate predictions for the test data. Let $y_i$ denote the observed value of the test data, $\hat{y}_i$ the predicted value, and $N$ the number of samples in the test data; the RMSE is given by
\begin{equation}
\mathrm{RMSE} = 
\sqrt{
\frac{1}{N}
\sum_{i=1}^{N}
\left( y_i - \hat{y}_i \right)^2}
\end{equation} 

Furthermore, regarding DTW, given two time-series $Y = (y_1, y_2, …, y_n)$ and $\hat{Y} = (\hat{y}_1, \hat{y}_2, …, \hat{y}_m)$, and a distance function $d(y_i, \hat{y}_j)$, the cumulative distance matrix $D(i, j)$ is defined as follows:
\begin{equation}
d(y_i, \hat{y}_j) = |y_i - \hat{y}_j|,
\end{equation} 
\begin{equation}
D(1,j) = d(y_1, \hat{y}_1),
\ \, \mathrm{with } \,\ j = 1,
\end{equation} 
\begin{equation}
D(1,j) = d(y_1, \hat{y}_j) + D(1,j-1),
\ \, \mathrm{with } \,\ 1 < j.
\end{equation} 
For $1 < i \le n$,
\begin{equation}
D(i,j) = d(y_i, \hat{y}_1) + D(i-1,1),
\ \, \mathrm{with } \, \, j = 1,
\end{equation} 
\begin{equation}
D(i,j) = d(y_i, \hat{y}_j) + \min\{D(i-1,j-1), D(i,j-1), D(i-1,j)\}
\ \, \mathrm{with } \, \, 1 < j.
\end{equation} 
Finally, the DTW distance between $Y$ and $\hat{Y}$ is defined as
\begin{equation}
\mathrm{DTW}(X,Y) = D(n,m).
\end{equation} 



\section{Implementation Details}
\label{environment}

All experiments are conducted on an Intel Xeon Platinum 8259CL CPU (4 vCPUs), 16 GB RAM, and one NVIDIA Tesla T4 GPU with 16 GB VRAM (CUDA 12.6).
The codebase is implemented in Python~3.12 using PyTorch~2.6.0 and TensorFlow~2.19.0.
A fixed random seed of \textbf{2025} is used throughout all experiments to ensure reproducibility.

\subsection{Dataset Details}
\label{sec:appendix_datasets}

Table~\ref{tab:dataset_details} summarises the datasets used in our experiments.
Each dataset is split chronologically into training, validation, and test sets
with a ratio of \textbf{6\,:\,2\,:\,2}.
The training split is z-score normalised (zero mean, unit variance), and the
same mean and standard deviation are applied to the validation and test splits.
To avoid information leakage at split boundaries, the validation and test
splits are extended backward by \texttt{input\_len} time steps so that the
first sliding window of each split can look back into the preceding partition.

The input length is fixed at $L_{\mathrm{input}}=12$ for all datasets.
Prediction lengths are chosen to reflect the seasonal periodicity of each
dataset: for the five univariate datasets we use
$L_{\mathrm{forecast}} \in \{3, 6, 9, 12\}$, while for ILI (weekly data with
an approximate 52-week annual cycle) we use
$L_{\mathrm{forecast}} \in \{2, 4, 8, 12\}$.
The seasonal period parameter (\texttt{window}) is set per-dataset based on
domain knowledge and is used by models that require explicit periodicity
information (e.g.\ S2IP-LLM, OLinear).

\begin{table}[ht]
\centering
\caption{Dataset statistics and experimental configuration.}
\label{tab:dataset_details}
\small
\begin{tabular}{lrcccc}
\hline
Dataset & Records & Frequency & Period & Batch size & $L_{\mathrm{forecast}}$ \\
\hline
Employees   & 427  & Monthly & 3  & 4 & \{3,6,9,12\} \\
Forest      & 517  & unknown & -  & 4 & \{3,6,9,12\} \\
German      & 221  & Every three months & 3 & 4 & \{3,6,9,12\} \\
ILI         & 966  & Weekly  & 2--4 & 4 & \{2,4,8,12\} \\
Inventories & 402  & Monthly & 3  & 4 & \{3,6,9,12\} \\
Consump.    & 96   & Yearly  & 3  & 4 & \{3,6,9,12\} \\
\hline
\end{tabular}
\end{table}

\subsection{Predictor Model Training}
\label{sec:appendix_predictor}

Table~\ref{tab:predictor_config} lists the hyperparameters shared across all
eight downstream predictor models.
All predictors are trained using AdamW with weight decay 0.1.
Early stopping monitors the validation MSE loss with a patience of 10 epochs.
The validation set used for early stopping is the standard validation split
(middle 20\% of the chronological data).

\begin{table}[ht]
\centering
\caption{Shared training configuration for all downstream TSF models.}
\label{tab:predictor_config}
\small
\begin{tabular}{lc}
\hline
Hyperparameter & Value \\
\hline
Optimizer & AdamW \\
Learning rate & $3 \times 10^{-4}$ \\
Weight decay & 0.1 \\
Max epochs & 100 \\
Early stopping patience & 10 epochs \\
Early stopping metric & Validation MSE \\
Batch size & 4 \\
Random seed & 2025 \\
\hline
\end{tabular}
\end{table}



\subsection{DAD4TS: Generator (Rectified-Flow)}
\label{sec:appendix_generator}

The generator in DAD4TS is based on a \textbf{Rectified Flow} formulation
with a 2D U-Net denoiser operating in a UMAP-derived latent space.
Table~\ref{tab:generator_config} summarises its configuration.

\paragraph{Conditioning via Mixture-of-Experts Embedding.}
The conditioning signal $\mathbf{c} \in \mathbb{R}^{512}$ is produced by a
\textbf{Top-1 Mixture-of-Experts (MoE)} gating network that selects among
three frozen time-series foundation model embedders:
\textbf{TiRex}, \textbf{Chronos-bolt-tiny}, and \textbf{Chronos-bolt-base}.
A learnable gating network (Linear$\to$ReLU$\to$Linear) takes the
normalised input sequence and outputs softmax scores over the three experts;
the expert with the highest score is selected for each sample.
Additionally, with 10\% probability during training, the conditioning vector
is replaced by a zero vector to enable classifier-free guidance (CFG) at
inference.

\begin{table}[ht]
\centering
\caption{DAD4TS Generator (Rectified-Flow Diffusion) configuration.}
\label{tab:generator_config}
\small
\begin{tabular}{lc}
\hline
Hyperparameter & Value \\
\hline
Optimizer & AdamW \\
Learning rate & $3 \times 10^{-3}$ \\
Weight decay & 0.1 \\
CFG dropout rate & 10\% \\
\hline
\multicolumn{2}{l}{\textit{2D U-Net Denoiser}} \\
\hline
Input dimension $D$ & $L_{\mathrm{input}}+L_{\mathrm{forecast}}$ \\
Base channels & 128 \\
Channel multipliers & (1, 2, 4) \\
Attention heads & 4 \\
Time embedding dim $H$ & 256 \\
Conditioning dim & 512 \\
GroupNorm groups & 8 \\
Activation & SiLU (ResBlock) / GEGLU (FFN) \\
\hline
\multicolumn{2}{l}{\textit{MoE Conditioning}} \\
\hline
Experts & TiRex, Chronos-tiny, Chronos-base \\
Gating & Top-1 hard routing \\
Expert output dim & 512 \\
\hline
\end{tabular}
\end{table}

\subsection{DAD4TS: Selector}
\label{sec:appendix_selector}

The Selector evaluates each generated sample and outputs a selection
probability $h \in [0,1]$.
It is a Transformer-based binary classifier that receives two types of
features: (1)~a temporal representation of the concatenated input--output
sequence, and (2)~statistical features including the sequence variance, mean,
and the predictor's per-sample loss on the generated data.

The Selector is trained via reinforcement learning method with a baseline
(exponential moving average of past rewards).
The reward signal is defined as the improvement in validation loss
(latter-half split) normalised by the selection ratio, encouraging the
Selector to pick samples that improve downstream prediction.
Table~\ref{tab:selector_config} details its architecture and training.

\begin{table}[ht]
\centering
\caption{Selector architecture Information.}
\label{tab:selector_config}
\small
\begin{tabular}{lc}
\hline
\multicolumn{2}{l}{\textit{Hyperparameter}} \\
\hline
Encoder & Transformer Encoder \\
$d_{\mathrm{model}}$ & 64 \\
\# of Attention heads & 4 \\
\# of Encoder layers & 2 \\
\# of FFN dim. & 128 \\
Dropout & 0.1 \\
Activation & GELU \\
Pooling & Adaptive average \\
Classifier & Linear($2 d_{\mathrm{model}} \to 64$)$\to$ReLU$\to$Linear($64 \to 1$) \\
Output & $\sigma(\mathrm{logit}/\tau)$, $\tau=0.7$ \\
\hline
\end{tabular}
\end{table}

\subsection{Baseline Generation Methods}
\label{sec:appendix_baselines}

Table~\ref{tab:gen_methods} summarises the key hyperparameters of each
data augmentation baseline.
All generation methods use the same random seed (2025) and are trained on the
training split only.
The number of generated samples equals the number of real training windows
(ratio 1:1).

\begin{table}[ht]
\centering
\caption{Hyperparameters of baseline data augmentation methods.}
\label{tab:gen_methods}
\small
\begin{tabular}{llc}
\hline
Method & Parameter & Value \\
\hline
\multirow{4}{*}{TimeGAN}
  & Module & GRU \\
  & Hidden dim / Layers & 24 / 3 \\
  & $\gamma$ & 1.0 \\
  & Epochs & 100 \\
\hline
TimeVAE
  & VAE type & TimeVAE (official) \\
\hline
\multirow{3}{*}{SMFG}
  & Image-based diffusion & 1000 timesteps \\
  & Loss type & Huber \\
  & Epochs & 100 \\
\hline
\multirow{7}{*}{Diffusion-TS}
  & Encoder / Decoder layers & 1 / 2 \\
  & $d_{\mathrm{model}}$ / heads & 64 / 4 \\
  & Sampling timesteps & 1000 \\
  & $\beta$ schedule & Cosine \\
  & Loss type & L1 \\
  & Learning rate & $3 \times 10^{-3}$ \\
  & Epochs & 100 \\
\hline
\multirow{3}{*}{DiffAT}
  & Iterations & 5000 \\
  & Diffusion $T$ & 200 \\
  & Residual layers & 10 \\
\hline
ReAugment
  & (default configuration) & -- \\
\hline
AutoDA-Timeseries
  & (default configuration) & -- \\
\hline
\end{tabular}
\end{table}

\newpage
\section{The DVRL-G Training Algorithm}
The learning algorithm of DVRL-G (Algorithm~\ref{alg:dvrlg_process}).

\label{DVRLG_Algorithm}

\begin{algorithm}
\caption{DVRL-G Process}
\label{alg:dvrlg_process}
\begin{algorithmic}[1]
\Require The datasets $\mathcal{D}_{train}$, $\mathcal{D}_{val}$ and $\mathcal{D}_{val}^{1/2}$, moving average of reward $\mathcal{R}_{ema}=0$, MSE loss function $\mathcal{L}$
\Models A Selector $\mathcal{S}_{\phi}$, time-series model $\mathcal{F}_{\theta}$ and diffusion model $\mathcal{M}_{\psi}$
\State \textbf{Initialize} $\phi, \theta$ and $\psi$
\Optimizers The optimizers $\mathcal{O}_\phi$, $\mathcal{O}_\theta$ and $\mathcal{O}_\psi$
\Hyperparameter Batch size $B$, number of epoch $N_{epoch}$, and number of mini-batches in $\mathcal{D}_{val}^{1/2}$ $Len_{\mathcal{D}_{val}^{1/2}}$
\Ensure Best predictor parameters $\phi, \theta$ and $\psi$

\For{each mini-batch $(\mathbf{x}_{train}, \mathbf{y}_{train}) \subset \mathcal{D}_{train}$}
    \State Compute prediction $\hat{\mathbf{y}}_{train} \gets \mathcal{F}_{\theta}(\mathbf{x}_{train})$
    \State Compute loss $\mathcal{L}_{train}(\hat{\mathbf{y}}_{train}, \mathbf{y}_{train})$
    \State Update $\theta$ using optimizer $\mathcal{O}_\theta$ with $\mathcal{L}_{train}$
\EndFor
\For{each mini-batch $(\mathbf{x}_{val^{1/2}}, \mathbf{y}_{val^{1/2}}) \subset \mathcal{D}_{val^{1/2}}$}
    \State Compute prediction $\hat{\mathbf{y}}_{val^{1/2}} \gets \mathcal{F}_{\theta}(\mathbf{x}_{val^{1/2}})$
    \State Compute loss $\mathcal{L}_{base}(\hat{\mathbf{y}}_{val^{1/2}}, \mathbf{y}_{val^{1/2}})$
\EndFor
\State Initialize parameters $\theta$ of $\mathcal{F}_{\theta}$
\\
\For{epoch $=1$ to $N_{epoch}$}
    \For{each mini-batch $(\mathbf{x}_{train}, \mathbf{y}_{train}) \subset \mathcal{D}_{train}$}
        \State \textbf{Train Diffusion Model} $\mathcal{M}_{\psi}$
        \Comment{Section \ref{sssec:diffusion_dad4ts} and Section \ref{sssec:training_dad4ts}}
        \State Update $\psi$ using optimizer $\mathcal{O}_\psi$ with MSE Loss
    \EndFor
    \State Initialize buffers $\mathcal{X}_f\gets[\ ],\ \mathcal{Y}_f\gets[\ ]$
    \For{each mini-batch $(\mathbf{x}_{train}, \mathbf{y}_{train}) \subset \mathcal{D}_{train}$}
        \State $x_{gen}, y_{gen} \gets \textbf{Data Generation}$
        \Comment{Section \ref{sssec:sampling_dad4ts}}
        \State add $x_{gen}$ and $y_{gen}$ to $\mathcal{X}_f,\mathcal{Y}_f$
    \EndFor
    \State Compute prediction $\hat{\mathcal{Y}_f} \gets \mathcal{F}_{\theta}(\mathcal{X}_f)$
    \State Compute loss $\mathcal{L}_{e}(\hat{\mathcal{Y}_f},\mathcal{Y}_f)$
    \For{each mini-batch $(\mathbf{x}_{train}, \mathbf{y}_{train}) \subset \mathcal{D}_{train}$}
    \State $p \gets \mathcal{S}_{\phi}(\mathcal{X}_f,\mathcal{Y}_f, L_{e})$
    \Comment{Section \ref{sssec:selector_model}}
    \State Sample mask $m \sim \mathrm{Bernoulli}(p)$
    \Comment{Section \ref{sssec:selector_model}}
    \State $(\mathcal{X}_s,\mathcal{Y}_s) \gets m \odot (\mathcal{X}_f,\mathcal{Y}_f)$
    \State Compute prediction $\hat{\mathcal{Y}_s} \gets \mathcal{F}_{\theta}(\mathcal{X}_s)$
    \State Compute loss $\mathcal{L}_{s}(\hat{\mathcal{Y}_s},\mathcal{Y}_s)$
    \State Update $\psi$ using optimizer $\mathcal{O}_\psi$ with $\mathcal{L}_{s}$
    \Comment{Section \ref{sssec:training_dad4ts}}
    \State Compute prediction $\hat{\mathbf{y}}_{train} \gets \mathcal{F}_{\theta}(\mathbf{x}_{train})$
    \State Compute loss $\mathcal{L}_{train}(\hat{\mathbf{y}}_{train},\mathbf{y}_{train})$
    \State $\tilde{\mathcal{L}}_{train} \gets \mathcal{L}_{train} + \mathcal{L}_{s}$
    \State Update $\theta$ using optimizer $\mathcal{O}_\theta$ with $\tilde{\mathcal{L}}_{train}$
    \State Initialize buffers $\mathcal{R}_c\gets[\ ]$
        \For{each mini-batch $(\mathbf{x}_{val^{1/2}}, \mathbf{y}_{val^{1/2}}) \subset \mathcal{D}_{val}^{1/2}$}
        \State Compute prediction $\hat{\mathbf{y}}_{val^{1/2}} \gets \mathcal{F}_{\theta}(\mathbf{x}_{val^{1/2}})$
        \State Compute loss $\mathcal{L}_{val^{1/2}}(\hat{\mathbf{y}}_{val^{1/2}},\mathbf{y}_{val^{1/2}})$
        \State Compute reward $\mathcal{R} \gets \mathcal{L}_{base}-\mathcal{L}_{val^{1/2}}$
        \State add $\mathcal{R}$ to $\mathcal{R}_c$
        \EndFor
    \State $\mathcal{R}_c \gets \frac{\mathcal{R}_c}{Len_{\mathcal{D}_{val}^{1/2}} \odot (\bar{m}+\epsilon)}$
    \State $\mathcal{R}_c \gets \mathcal{R}_c-\mathcal{R}_{ema}$
    \State $\mathcal{R}_{ema} \gets 0.9  \mathcal{R}_{ema} + 0.1 \mathcal{R}_c$
    \State Compute per sample quality $\mathcal{Q} \gets \mathrm{Standardize}(\mathcal{R}_c \odot -(L_{e}))$
    \State Compute log-probability $\log \Pi_\phi(m|p) \gets m\odot \log(p+\epsilon) + (1-m)\odot \log(1-p+\epsilon)$
    \State Compute final reward $\mathcal{R}_{\phi} \gets -(\mathcal{Q}\odot\log \Pi_\phi(m|p))$
    \State Update $\phi$ using optimizer $\mathcal{O}_\phi$ with $\mathcal{R}_{\phi}$
    \EndFor
\EndFor

\end{algorithmic}
\end{algorithm}

\newpage
\section{All results for DAD4TS}
Table~\ref{employees-DAD4TS}, ~\ref{Forest-DAD4TS}, ~\ref{German-DAD4TS}, ~\ref{ILI-DAD4TS}, ~\ref{Inventories-DAD4TS} and ~\ref{Consump-DAD4TS} show the results for each prediction length in terms of RMSE and the DTW metric for DAD4TS.

\begin{table}[H]
\centering
\caption{== Employees == RMSE and DTW results for each forecast lengths.}
\label{employees-DAD4TS}
\resizebox{\textwidth}{!}{
\begin{tabular}{lcccccccc}
\toprule
Model & \multicolumn{2}{c}{3} & \multicolumn{2}{c}{6} & \multicolumn{2}{c}{9} & \multicolumn{2}{c}{12} \\
\cmidrule(lr){2-3} \cmidrule(lr){4-5} \cmidrule(lr){6-7} \cmidrule(lr){8-9} 
& \multicolumn{1}{c}{RMSE} & \multicolumn{1}{c}{DTW} & \multicolumn{1}{c}{RMSE} & \multicolumn{1}{c}{DTW} & \multicolumn{1}{c}{RMSE} & \multicolumn{1}{c}{DTW} & \multicolumn{1}{c}{RMSE} & \multicolumn{1}{c}{DTW} \\
\midrule
RNN & 0.467 $\pm$ 0.315 & 1.4 $\pm$ 0.947 & 0.487 $\pm$ 0.326 & 2.9 $\pm$ 1.97 & 0.405 $\pm$ 0.262 & 3.56 $\pm$ 2.38 & 0.457 $\pm$ 0.276 & 5.3 $\pm$ 3.4 \\
LSTM & 0.293 $\pm$ 0.245 & 0.869 $\pm$ 0.731 & 0.349 $\pm$ 0.24 & 2.06 $\pm$ 1.44 & 0.417 $\pm$ 0.288 & 3.68 $\pm$ 2.62 & 0.443 $\pm$ 0.293 & 5.14 $\pm$ 3.62 \\
Transformer & 0.557 $\pm$ 0.488 & 1.66 $\pm$ 1.47 & 0.537 $\pm$ 0.447 & 3.19 $\pm$ 2.7 & 0.51 $\pm$ 0.38 & 4.51 $\pm$ 3.46 & 0.558 $\pm$ 0.388 & 6.56 $\pm$ 4.72 \\
iTransformer & 0.0795 $\pm$ 0.157 & 0.21 $\pm$ 0.421 & 0.122 $\pm$ 0.185 & 0.562 $\pm$ 0.92 & 0.165 $\pm$ 0.226 & 1.21 $\pm$ 1.85 & 0.209 $\pm$ 0.241 & 1.96 $\pm$ 2.66 \\
PatchTST & 0.0814 $\pm$ 0.144 & 0.212 $\pm$ 0.382 & 0.124 $\pm$ 0.177 & 0.599 $\pm$ 0.936 & 0.161 $\pm$ 0.199 & 1.11 $\pm$ 1.61 & 0.205 $\pm$ 0.204 & 1.81 $\pm$ 2.27 \\
TimesNet & 0.087 $\pm$ 0.164 & 0.229 $\pm$ 0.438 & 0.126 $\pm$ 0.194 & 0.629 $\pm$ 1.05 & 0.161 $\pm$ 0.216 & 1.17 $\pm$ 1.74 & 0.203 $\pm$ 0.23 & 1.93 $\pm$ 2.53 \\
S2IP-LLM & 0.0958 $\pm$ 0.172 & 0.261 $\pm$ 0.476 & 0.12 $\pm$ 0.178 & 0.575 $\pm$ 0.928 & 0.169 $\pm$ 0.233 & 1.22 $\pm$ 1.89 & 0.229 $\pm$ 0.285 & 2.18 $\pm$ 3.2 \\
OLinear & 0.0933 $\pm$ 0.166 & 0.239 $\pm$ 0.432 & 0.121 $\pm$ 0.177 & 0.583 $\pm$ 0.932 & 0.15 $\pm$ 0.19 & 1.07 $\pm$ 1.5 & 0.205 $\pm$ 0.201 & 1.87 $\pm$ 2.24 \\
\bottomrule
\end{tabular}
}
\end{table}

\begin{table}[H]
\centering
\caption{== Forest == RMSE and DTW results for each forecast lengths.}
\label{Forest-DAD4TS}
\resizebox{\textwidth}{!}{
\begin{tabular}{lcccccccc}
\toprule
Model & \multicolumn{2}{c}{3} & \multicolumn{2}{c}{6} & \multicolumn{2}{c}{9} & \multicolumn{2}{c}{12} \\
\cmidrule(lr){2-3} \cmidrule(lr){4-5} \cmidrule(lr){6-7} \cmidrule(lr){8-9} 
& \multicolumn{1}{c}{RMSE} & \multicolumn{1}{c}{DTW} & \multicolumn{1}{c}{RMSE} & \multicolumn{1}{c}{DTW} & \multicolumn{1}{c}{RMSE} & \multicolumn{1}{c}{DTW} & \multicolumn{1}{c}{RMSE} & \multicolumn{1}{c}{DTW} \\
\midrule
RNN & 0.895 $\pm$ 0.436 & 2.43 $\pm$ 1.27 & 1.05 $\pm$ 0.422 & 5.33 $\pm$ 2.47 & 1.05 $\pm$ 0.352 & 7.56 $\pm$ 3.05 & 1.22 $\pm$ 0.348 & 11.3 $\pm$ 4.1 \\
LSTM & 0.916 $\pm$ 0.473 & 2.46 $\pm$ 1.37 & 1.06 $\pm$ 0.449 & 5.4 $\pm$ 2.66 & 1.1 $\pm$ 0.394 & 8.06 $\pm$ 3.46 & 1.15 $\pm$ 0.353 & 11.4 $\pm$ 4.04 \\
Transformer & 0.959 $\pm$ 0.508 & 2.61 $\pm$ 1.5 & 1.05 $\pm$ 0.403 & 5.52 $\pm$ 2.32 & 1.19 $\pm$ 0.416 & 8.95 $\pm$ 3.62 & 1.22 $\pm$ 0.36 & 12.2 $\pm$ 4.16 \\
iTransformer & 0.916 $\pm$ 0.424 & 2.4 $\pm$ 1.22 & 0.999 $\pm$ 0.36 & 4.87 $\pm$ 2.14 & 1.07 $\pm$ 0.356 & 7.58 $\pm$ 3.13 & 1.12 $\pm$ 0.339 & 10.1 $\pm$ 3.78 \\
PatchTST & 0.977 $\pm$ 0.408 & 2.57 $\pm$ 1.19 & 1.09 $\pm$ 0.387 & 5.24 $\pm$ 2.32 & 1.12 $\pm$ 0.435 & 7.76 $\pm$ 3.95 & 1.21 $\pm$ 0.407 & 10.5 $\pm$ 4.34 \\
TimesNet & 0.929 $\pm$ 0.432 & 2.39 $\pm$ 1.25 & 1.08 $\pm$ 0.394 & 5.07 $\pm$ 2.28 & 1.2 $\pm$ 0.436 & 8.04 $\pm$ 3.86 & 1.16 $\pm$ 0.372 & 10.1 $\pm$ 3.84 \\
S2IP-LLM & 0.929 $\pm$ 0.401 & 2.49 $\pm$ 1.15 & 1.03 $\pm$ 0.38 & 4.81 $\pm$ 2.3 & 1.06 $\pm$ 0.368 & 7.43 $\pm$ 3.22 & 1.15 $\pm$ 0.374 & 10.1 $\pm$ 4.3 \\
OLinear & 0.921 $\pm$ 0.409 & 2.48 $\pm$ 1.16 & 0.998 $\pm$ 0.397 & 4.97 $\pm$ 2.36 & 1.11 $\pm$ 0.384 & 7.77 $\pm$ 3.32 & 1.18 $\pm$ 0.403 & 10.6 $\pm$ 4.8 \\
\bottomrule
\end{tabular}
}
\end{table}

\begin{table}[H]
\centering
\caption{== German == RMSE and DTW results for each forecast lengths.}
\label{German-DAD4TS}
\resizebox{\textwidth}{!}{
\begin{tabular}{lcccccccc}
\toprule
Model & \multicolumn{2}{c}{3} & \multicolumn{2}{c}{6} & \multicolumn{2}{c}{9} & \multicolumn{2}{c}{12} \\
\cmidrule(lr){2-3} \cmidrule(lr){4-5} \cmidrule(lr){6-7} \cmidrule(lr){8-9} 
& \multicolumn{1}{c}{RMSE} & \multicolumn{1}{c}{DTW} & \multicolumn{1}{c}{RMSE} & \multicolumn{1}{c}{DTW} & \multicolumn{1}{c}{RMSE} & \multicolumn{1}{c}{DTW} & \multicolumn{1}{c}{RMSE} & \multicolumn{1}{c}{DTW} \\
\midrule
RNN & 0.631 $\pm$ 0.445 & 1.71 $\pm$ 1.33 & 0.952 $\pm$ 0.525 & 4.96 $\pm$ 3.07 & 1.31 $\pm$ 0.427 & 10.1 $\pm$ 3.85 & 1.48 $\pm$ 0.401 & 14 $\pm$ 4.62 \\
LSTM & 0.895 $\pm$ 0.509 & 2.53 $\pm$ 1.56 & 1.34 $\pm$ 0.478 & 7.28 $\pm$ 3 & 1.51 $\pm$ 0.378 & 11.6 $\pm$ 3.36 & 1.69 $\pm$ 0.365 & 16.8 $\pm$ 3.86 \\
Transformer & 0.737 $\pm$ 0.462 & 2.07 $\pm$ 1.38 & 1.3 $\pm$ 0.49 & 7.15 $\pm$ 2.84 & 1.57 $\pm$ 0.439 & 12.8 $\pm$ 3.81 & 1.74 $\pm$ 0.495 & 17.5 $\pm$ 5.08 \\
iTransformer & 0.59 $\pm$ 0.538 & 1.58 $\pm$ 1.52 & 1.08 $\pm$ 0.84 & 5.35 $\pm$ 4.68 & 1.61 $\pm$ 0.954 & 12.1 $\pm$ 8.11 & 2.02 $\pm$ 0.812 & 19.9 $\pm$ 8.81 \\
PatchTST & 0.533 $\pm$ 0.571 & 1.39 $\pm$ 1.53 & 1.12 $\pm$ 0.76 & 5.52 $\pm$ 4.04 & 1.73 $\pm$ 0.832 & 13.2 $\pm$ 6.84 & 2.07 $\pm$ 0.725 & 19.4 $\pm$ 6.72 \\
TimesNet & 0.567 $\pm$ 0.588 & 1.51 $\pm$ 1.65 & 1.11 $\pm$ 0.714 & 5.67 $\pm$ 3.84 & 1.6 $\pm$ 0.974 & 11.9 $\pm$ 7.99 & 2.02 $\pm$ 0.838 & 19.7 $\pm$ 9 \\
S2IP-LLM & 0.564 $\pm$ 0.479 & 1.45 $\pm$ 1.32 & 1.2 $\pm$ 0.733 & 6.11 $\pm$ 4.08 & 1.58 $\pm$ 1.21 & 11.4 $\pm$ 10.4 & 2.17 $\pm$ 0.547 & 22.5 $\pm$ 5.26 \\
OLinear & 0.761 $\pm$ 0.446 & 2.07 $\pm$ 1.27 & 1.29 $\pm$ 0.541 & 7 $\pm$ 2.88 & 1.63 $\pm$ 0.961 & 11.8 $\pm$ 8.2 & 2.03 $\pm$ 0.785 & 19.8 $\pm$ 8.15 \\
\bottomrule
\end{tabular}
}
\end{table}

\begin{table}[H]
\centering
\caption{== ILI == RMSE and DTW results for each forecast lengths.}
\label{ILI-DAD4TS}
\resizebox{\textwidth}{!}{
\begin{tabular}{lcccccccc}
\toprule
Model & \multicolumn{2}{c}{2} & \multicolumn{2}{c}{4} & \multicolumn{2}{c}{8} & \multicolumn{2}{c}{12} \\
\cmidrule(lr){2-3} \cmidrule(lr){4-5} \cmidrule(lr){6-7} \cmidrule(lr){8-9} 
& \multicolumn{1}{c}{RMSE} & \multicolumn{1}{c}{DTW} & \multicolumn{1}{c}{RMSE} & \multicolumn{1}{c}{DTW} & \multicolumn{1}{c}{RMSE} & \multicolumn{1}{c}{DTW} & \multicolumn{1}{c}{RMSE} & \multicolumn{1}{c}{DTW} \\
\midrule
RNN & 0.866 $\pm$ 0.678 & 1.69 $\pm$ 1.36 & 1.19 $\pm$ 0.824 & 4.61 $\pm$ 3.33 & 1.14 $\pm$ 0.751 & 8.53 $\pm$ 6.14 & 1.31 $\pm$ 0.72 & 14.7 $\pm$ 8.75 \\
LSTM & 0.861 $\pm$ 0.672 & 1.69 $\pm$ 1.35 & 0.931 $\pm$ 0.653 & 3.58 $\pm$ 2.63 & 1.11 $\pm$ 0.69 & 8.36 $\pm$ 5.57 & 1.36 $\pm$ 0.789 & 14.9 $\pm$ 9.79 \\
Transformer & 0.957 $\pm$ 0.801 & 1.88 $\pm$ 1.61 & 0.995 $\pm$ 0.752 & 3.83 $\pm$ 3.03 & 1.19 $\pm$ 0.789 & 9.04 $\pm$ 6.41 & 1.33 $\pm$ 0.754 & 14.9 $\pm$ 9.3 \\
iTransformer & 0.296 $\pm$ 0.306 & 0.542 $\pm$ 0.579 & 0.424 $\pm$ 0.382 & 1.46 $\pm$ 1.42 & 0.622 $\pm$ 0.596 & 4.05 $\pm$ 4.33 & 0.815 $\pm$ 0.742 & 7.75 $\pm$ 7.79 \\
PatchTST & 0.28 $\pm$ 0.289 & 0.51 $\pm$ 0.55 & 0.395 $\pm$ 0.375 & 1.35 $\pm$ 1.38 & 0.677 $\pm$ 0.602 & 4.37 $\pm$ 4.31 & 0.788 $\pm$ 0.643 & 7.35 $\pm$ 6.64 \\
TimesNet & 0.315 $\pm$ 0.324 & 0.585 $\pm$ 0.621 & 0.418 $\pm$ 0.384 & 1.42 $\pm$ 1.43 & 0.649 $\pm$ 0.535 & 4.15 $\pm$ 3.9 & 0.755 $\pm$ 0.586 & 6.89 $\pm$ 6.03 \\
S2IP-LLM & 0.306 $\pm$ 0.293 & 0.563 $\pm$ 0.551 & 0.454 $\pm$ 0.445 & 1.56 $\pm$ 1.64 & 0.661 $\pm$ 0.609 & 4.25 $\pm$ 4.32 & 0.878 $\pm$ 0.831 & 8.27 $\pm$ 8.52 \\
OLinear & 0.304 $\pm$ 0.323 & 0.557 $\pm$ 0.61 & 0.415 $\pm$ 0.402 & 1.43 $\pm$ 1.49 & 0.652 $\pm$ 0.576 & 4.31 $\pm$ 4.21 & 0.772 $\pm$ 0.673 & 7.42 $\pm$ 7.22 \\
\bottomrule
\end{tabular}
}
\end{table}

\begin{table}[H]
\centering
\caption{== Inventories == RMSE and DTW results for each forecast lengths.}
\label{Inventories-DAD4TS}
\resizebox{\textwidth}{!}{
\begin{tabular}{lcccccccc}
\toprule
Model & \multicolumn{2}{c}{3} & \multicolumn{2}{c}{6} & \multicolumn{2}{c}{9} & \multicolumn{2}{c}{12} \\
\cmidrule(lr){2-3} \cmidrule(lr){4-5} \cmidrule(lr){6-7} \cmidrule(lr){8-9} 
& \multicolumn{1}{c}{RMSE} & \multicolumn{1}{c}{DTW} & \multicolumn{1}{c}{RMSE} & \multicolumn{1}{c}{DTW} & \multicolumn{1}{c}{RMSE} & \multicolumn{1}{c}{DTW} & \multicolumn{1}{c}{RMSE} & \multicolumn{1}{c}{DTW} \\
\midrule
RNN & 0.286 $\pm$ 0.411 & 0.765 $\pm$ 1.07 & 0.415 $\pm$ 0.448 & 2.06 $\pm$ 2.06 & 0.521 $\pm$ 0.489 & 3.73 $\pm$ 3.26 & 0.654 $\pm$ 0.527 & 6.38 $\pm$ 4.7 \\
LSTM & 0.312 $\pm$ 0.437 & 0.841 $\pm$ 1.16 & 0.42 $\pm$ 0.473 & 2.1 $\pm$ 2.26 & 0.546 $\pm$ 0.499 & 4.01 $\pm$ 3.43 & 0.91 $\pm$ 0.634 & 9.84 $\pm$ 6.5 \\
Transformer & 0.331 $\pm$ 0.469 & 0.903 $\pm$ 1.25 & 0.438 $\pm$ 0.516 & 2.25 $\pm$ 2.53 & 0.595 $\pm$ 0.553 & 4.54 $\pm$ 4.01 & 0.668 $\pm$ 0.585 & 6.57 $\pm$ 5.53 \\
iTransformer & 0.351 $\pm$ 0.512 & 0.949 $\pm$ 1.37 & 0.5 $\pm$ 0.554 & 2.56 $\pm$ 2.83 & 0.637 $\pm$ 0.592 & 4.71 $\pm$ 4.35 & 0.722 $\pm$ 0.618 & 6.87 $\pm$ 5.92 \\
PatchTST & 0.357 $\pm$ 0.499 & 0.975 $\pm$ 1.36 & 0.52 $\pm$ 0.579 & 2.62 $\pm$ 2.93 & 0.632 $\pm$ 0.594 & 4.61 $\pm$ 4.29 & 0.753 $\pm$ 0.598 & 7.03 $\pm$ 5.34 \\
TimesNet & 0.389 $\pm$ 0.541 & 1.04 $\pm$ 1.47 & 0.512 $\pm$ 0.567 & 2.52 $\pm$ 2.8 & 0.646 $\pm$ 0.608 & 4.71 $\pm$ 4.45 & 0.795 $\pm$ 0.64 & 7.6 $\pm$ 6.16 \\
S2IP-LLM & 0.336 $\pm$ 0.514 & 0.89 $\pm$ 1.38 & 0.477 $\pm$ 0.594 & 2.37 $\pm$ 3.08 & 0.654 $\pm$ 0.615 & 4.9 $\pm$ 4.7 & 0.739 $\pm$ 0.598 & 6.95 $\pm$ 5.29 \\
OLinear & 0.342 $\pm$ 0.475 & 0.921 $\pm$ 1.28 & 0.493 $\pm$ 0.566 & 2.49 $\pm$ 2.89 & 0.652 $\pm$ 0.596 & 4.9 $\pm$ 4.38 & 0.752 $\pm$ 0.601 & 7.37 $\pm$ 5.84 \\
\bottomrule
\end{tabular}
}
\end{table}

\begin{table}[H]
\centering
\caption{== Consump == RMSE and DTW results for each forecast lengths.}
\label{Consump-DAD4TS}
\resizebox{\textwidth}{!}{
\begin{tabular}{lcccccccc}
\toprule
Model & \multicolumn{2}{c}{3} & \multicolumn{2}{c}{6} & \multicolumn{2}{c}{9} & \multicolumn{2}{c}{12} \\
\cmidrule(lr){2-3} \cmidrule(lr){4-5} \cmidrule(lr){6-7} \cmidrule(lr){8-9} 
& \multicolumn{1}{c}{RMSE} & \multicolumn{1}{c}{DTW} & \multicolumn{1}{c}{RMSE} & \multicolumn{1}{c}{DTW} & \multicolumn{1}{c}{RMSE} & \multicolumn{1}{c}{DTW} & \multicolumn{1}{c}{RMSE} & \multicolumn{1}{c}{DTW} \\
\midrule
RNN & 4.19 $\pm$ 0.807 & 12.6 $\pm$ 2.42 & 4.25 $\pm$ 0.614 & 25.5 $\pm$ 3.66 & 5.85 $\pm$ 0.494 & 52.6 $\pm$ 4.38 & 6.07 $\pm$ 0.394 & 72.6 $\pm$ 4.62 \\
LSTM & 5.52 $\pm$ 0.812 & 16.6 $\pm$ 2.43 & 5.67 $\pm$ 0.627 & 34 $\pm$ 3.73 & 5.75 $\pm$ 0.494 & 51.6 $\pm$ 4.38 & 5.89 $\pm$ 0.394 & 70.4 $\pm$ 4.62 \\
Transformer & 4.29 $\pm$ 0.81 & 12.9 $\pm$ 2.42 & 5.68 $\pm$ 0.629 & 34.1 $\pm$ 3.74 & 5.82 $\pm$ 0.497 & 52.3 $\pm$ 4.41 & 5.86 $\pm$ 0.394 & 70.2 $\pm$ 4.63 \\
iTransformer & 0.281 $\pm$ 0.169 & 0.755 $\pm$ 0.483 & 0.388 $\pm$ 0.182 & 1.6 $\pm$ 0.862 & 0.377 $\pm$ 0.157 & 1.75 $\pm$ 0.617 & 0.329 $\pm$ 0.125 & 1.51 $\pm$ 0.374 \\
PatchTST & 0.204 $\pm$ 0.221 & 0.531 $\pm$ 0.613 & 0.356 $\pm$ 0.186 & 1.37 $\pm$ 0.829 & 0.375 $\pm$ 0.176 & 2.03 $\pm$ 0.998 & 0.806 $\pm$ 0.355 & 4.94 $\pm$ 3.11 \\
TimesNet & 0.273 $\pm$ 0.174 & 0.71 $\pm$ 0.494 & 0.364 $\pm$ 0.147 & 1.41 $\pm$ 0.772 & 0.278 $\pm$ 0.0997 & 1.36 $\pm$ 0.518 & 0.377 $\pm$ 0.119 & 2.07 $\pm$ 0.874 \\
S2IP-LLM & 0.279 $\pm$ 0.196 & 0.762 $\pm$ 0.566 & 0.432 $\pm$ 0.173 & 1.73 $\pm$ 0.849 & 0.46 $\pm$ 0.23 & 2.46 $\pm$ 1.39 & 0.547 $\pm$ 0.167 & 4.1 $\pm$ 1.09 \\
OLinear & 0.253 $\pm$ 0.244 & 0.669 $\pm$ 0.696 & 0.328 $\pm$ 0.235 & 1.27 $\pm$ 0.991 & 0.415 $\pm$ 0.249 & 2.22 $\pm$ 1.57 & 1.22 $\pm$ 0.549 & 8.28 $\pm$ 5.55 \\
\bottomrule
\end{tabular}
}
\end{table}

\newpage
\section{Visualization of Generated Data Using Various Methods}
\begin{figure}
    \centering
    \includegraphics[width=0.7\linewidth]{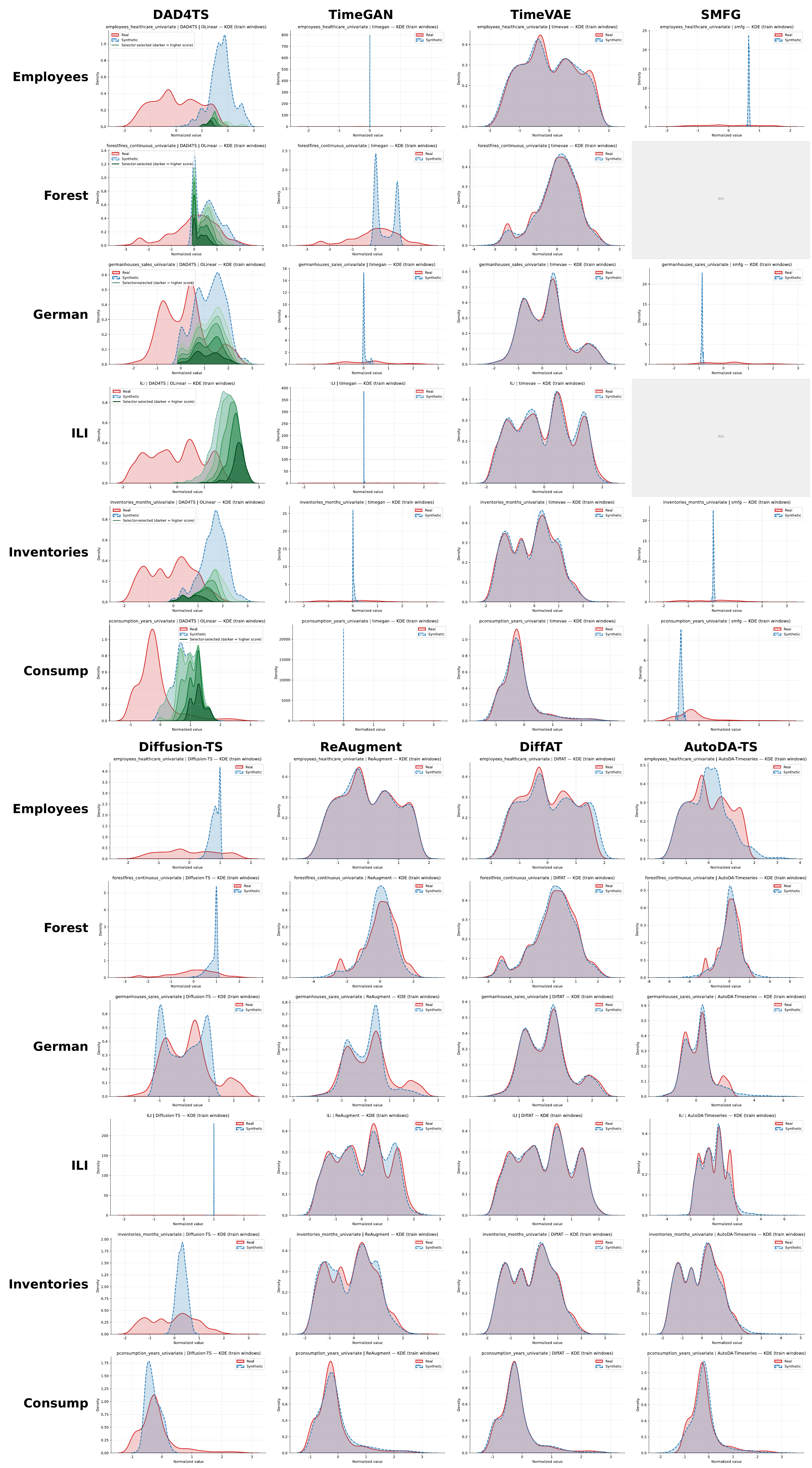}
    \caption{The KDE-distribution of data generated by the proposed method and the methods proposed in previous studies. Red indicates real data, and blue indicates generated data. Green indicates generated data selected by the Selector as suitable for training. The darker the color, the higher the data is rated as valuable for learning. In this figure, the German dataset is used, and the input and forecast lengths are fixed at 12. “N/A” indicates that data could not be generated in the experimental environment due to computational costs.}
    \label{compare_distribution_kde}
\end{figure}

\begin{figure}
    \centering
    \includegraphics[width=0.6\linewidth]{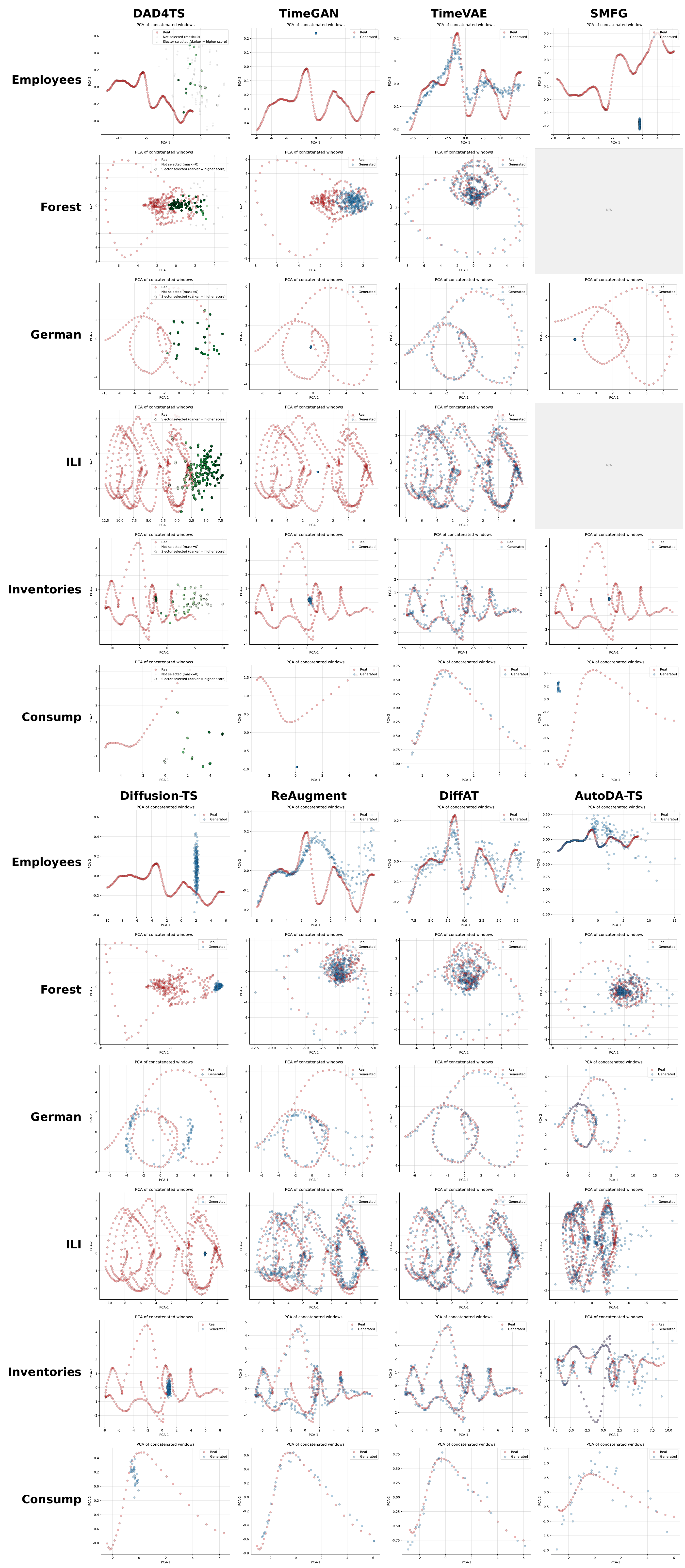}
    \caption{The PCA-distribution of data generated by the proposed method and the methods proposed in previous studies. Red indicates real data, and blue indicates generated data. Green indicates generated data selected by the Selector as suitable for training. The darker the color, the higher the data is rated as valuable for learning. In this figure, the German dataset is used, and the input and forecast lengths are fixed at 12. “N/A” indicates that data could not be generated in the experimental environment due to computational costs.}
    \label{compare_distribution_pca}
\end{figure}

\begin{figure}
    \centering
    \includegraphics[width=1.0\linewidth]{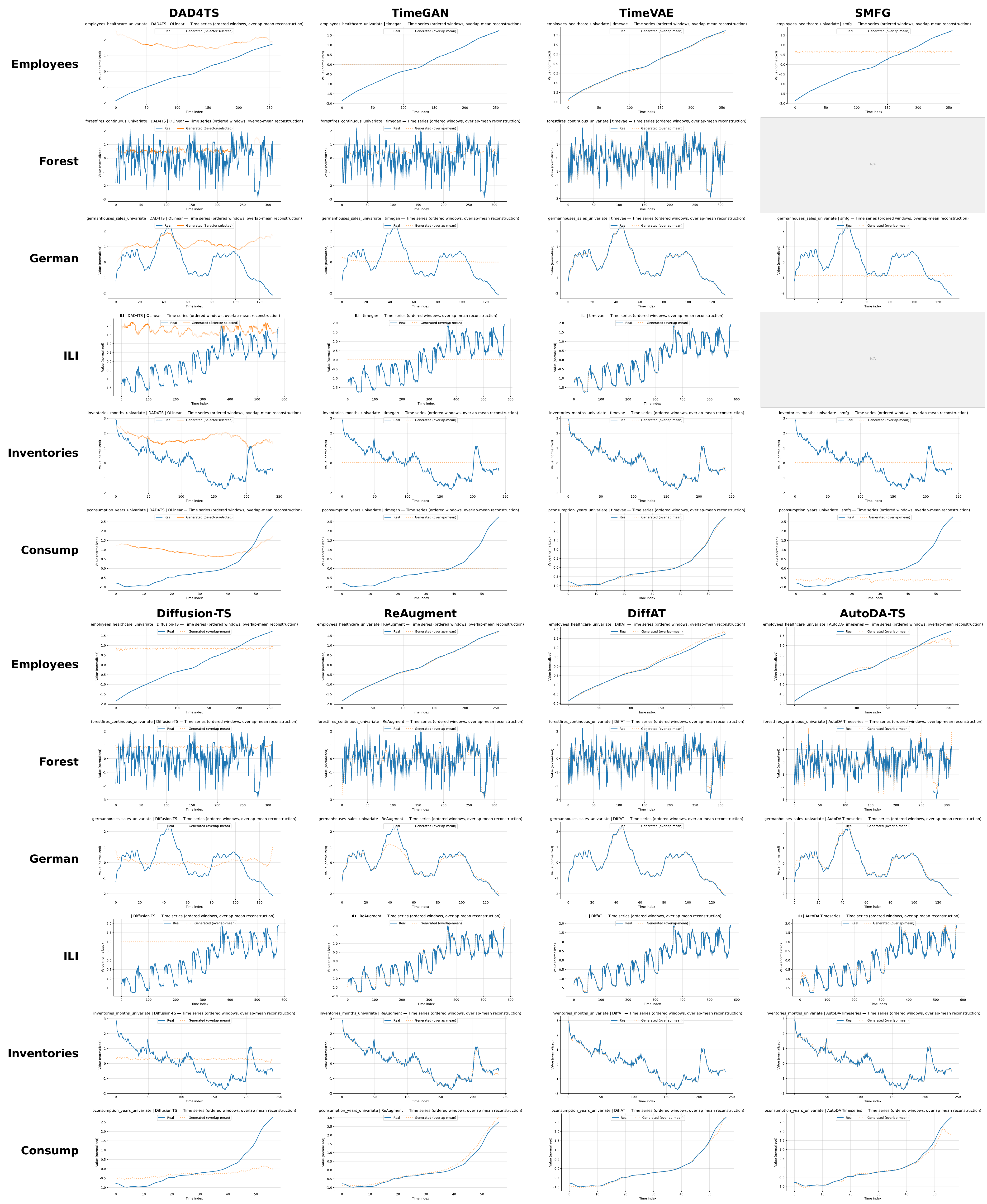}
    \caption{Visualization of the generated data from a time-series perspective. For DAD4TS, darker colors indicate data with higher utility as training data. In this figure, the German dataset is used, and the input and forecast lengths are fixed at 12. “N/A” indicates that data could not be generated in the experimental environment due to computational costs.}
    \label{compare_distribution_series}
\end{figure}

The following results show the visualization of data generated using the DAD4TS and the methods employed in previous studies. The figure are Figure~\ref{compare_distribution_kde}, ~\ref{compare_distribution_pca} and ~\ref{compare_distribution_series}. The visualization results for the DAD4TS are based on those obtained using OLInear for the time series forecasting model.

\end{document}